\documentclass[letterpaper]{article} 
\usepackage{aaai24}  
\usepackage{times}  
\usepackage{helvet}  
\usepackage{courier}  
\usepackage[hyphens]{url}  
\usepackage{graphicx} 
\usepackage{natbib}  
\usepackage{caption} 
\usepackage{algorithm}
\usepackage{algorithmic}

\usepackage{newfloat}
\usepackage{listings}
\DeclareCaptionStyle{ruled}{labelfont=normalfont,labelsep=colon,strut=off} 
\floatstyle{ruled}
\newfloat{listing}{tb}{lst}{}
\floatname{listing}{Listing}
\usepackage{amsmath}
\usepackage{amssymb}
\usepackage{mathtools}
\usepackage{amsthm}
\usepackage{bm}
\usepackage{amsfonts}  
\usepackage{subcaption}
\usepackage{multirow}  
\usepackage{makecell}  
\usepackage{textcomp}  
\usepackage{numprint}  
\usepackage{url}            
\usepackage{booktabs}       
\usepackage[table,x11names,dvipsnames]{xcolor}         
\npthousandsep{,}
\usepackage{soul}

\usepackage{etoolbox}
\usepackage{tikz}
\usetikzlibrary{calc,shapes.geometric,trees,positioning,arrows,arrows.meta,backgrounds,fit,decorations.pathreplacing,decorations.markings,calligraphy,matrix}

\makeatletter
\DeclareRobustCommand{\rvdots}{%
  \vbox{
    \baselineskip4\p@\lineskiplimit\z@
    \kern-\p@
    \hbox{.}\hbox{.}\hbox{.}
  }%
}
\makeatother

\newcommand{\unit}{0.4cm}

\definecolor{myred}{HTML}{FC8D59}
\definecolor{myyellow}{HTML}{FFFFBF}
\definecolor{myblue}{HTML}{91BFDB}

\definecolor{c0}{HTML}{9b9b7a}
\definecolor{c1}{HTML}{baa587}
\definecolor{c2}{HTML}{e8ac65}
\definecolor{c3}{HTML}{d08c60}

\title{One Self-Configurable Model to Solve Many Abstract Visual Reasoning Problems}
\author{%
  Mikołaj Małkiński\textsuperscript{\rm 1},
  Jacek Mańdziuk\textsuperscript{\rm 1,\rm 2}
}
\affiliations{%
    \textsuperscript{\rm 1}Faculty of Mathematics and Information Science, Warsaw University of Technology, Warsaw, Poland\\
    \textsuperscript{\rm 2}Faculty of Computer Science, AGH University of Krakow, Krakow, Poland\\
    mikolaj.malkinski.dokt@pw.edu.pl, mandziuk@mini.pw.edu.pl
}

\begin{document}

\maketitle

\begin{abstract}
Abstract Visual Reasoning (AVR) comprises a wide selection of various problems similar to those used in human IQ tests.
Recent years have brought dynamic progress in solving particular AVR tasks, however, in the contemporary literature AVR problems are largely dealt with in isolation, leading to highly specialized task-specific methods.
With the aim of developing universal learning systems in the AVR domain, we propose the \emph{unified model for solving Single-Choice Abstract visual Reasoning tasks} (SCAR), capable of solving various single-choice AVR tasks, without making any a priori assumptions about the task structure, in particular the number and location of panels.
The proposed model relies on a novel \textit{Structure-Aware dynamic Layer} (SAL), which adapts its weights to the structure of the considered AVR problem.
Experiments conducted on Raven's Progressive Matrices, Visual Analogy Problems, and Odd One Out problems show that SCAR (SAL-based models, in general) effectively solves diverse AVR tasks, and its performance is on par with the state-of-the-art task-specific baselines.
What is more, SCAR demonstrates effective knowledge reuse in multi-task and transfer learning settings.
To our knowledge, this work is the first successful attempt to construct a general single-choice AVR solver relying on self-configurable architecture and unified solving method.
With this work we aim to stimulate and foster progress on task-independent research paths in the AVR domain, with the long-term goal of development of a general AVR solver.
\end{abstract}

\section{Introduction}\label{sec:introduction}

For many years, Abstract Visual Reasoning (AVR) has been a highly challenging area of artificial intelligence (AI) research~\cite{hernandez2016computer}.
In a typical AVR task, the solver has to identify a set of rules (visual patterns) that govern the distribution of a set of 2D shapes with certain attributes.
The shapes are scattered across several \textit{image panels}, which are arranged in a specific 2D structure.
Popularised by human IQ tests, the prevailing form of task representation is a single-choice matrix, where a test-taker has to select one of the panels as a task answer.
A common example are Raven's Progressive Matrices (RPMs)~\cite{raven1936mental,raven1998raven} that consist of a $3\times3$ grid of image panels with the bottom-right panel being missing.
The task is to complete the grid with one of the panels picked from a separate set of \textit{answer panels} that best fits the arrangement of the image panels.
Top-left part of Fig.~\ref{fig:intro} presents an RPM with $9$ image panels and $8$ answer panels (A -- H).
Additional examples are presented in Appendix~\ref{sec:examples}.

\definecolor{rpm}{HTML}{FFFFBF}
\definecolor{vap}{HTML}{91BFDB}
\definecolor{ooo}{HTML}{5E35B1}
\definecolor{layer}{HTML}{E1E1E1}
\definecolor{activation}{HTML}{A9C8C0}

\tikzstyle{emptylayer} = [rectangle, inner sep=0, minimum height=1.2*\unit, minimum width=4*\unit]
\tikzstyle{layer} = [emptylayer, rounded corners, text centered, draw=black, fill=layer, inner sep=2pt, minimum width=6*\unit]
\tikzstyle{background} = [draw, fill=black!5, rounded corners=5pt, densely dashed, inner sep=0]
\tikzstyle{panel} = [thick, rectangle, draw=black, inner sep=0]
\tikzstyle{label} = [rectangle, inner sep=0]
\tikzstyle{title} = [align=center, inner sep=2pt, font={\footnotesize}]

\tikzstyle{embedding} = [circle, text centered, draw=black, fill=layer!50, inner sep=0, minimum height=1.2*\unit, minimum width=1.2*\unit]

\tikzstyle{arrow} = [->,>={Latex[scale=1]}]

\newcommand{\rpm}[1]{\includegraphics[width=0.04\textwidth]{images/rpm/context_#1}}
\newcommand{\rpmans}[1]{\includegraphics[width=0.04\textwidth]{images/rpm/answer_#1}}
\newcommand{\vap}[1]{\includegraphics[width=0.04\textwidth]{images/vap/context_#1}}
\newcommand{\vapans}[1]{\includegraphics[width=0.04\textwidth]{images/vap/answer_#1}}
\newcommand{\ooo}[1]{\includegraphics[width=0.04\textwidth]{images/ooo/context_#1}}

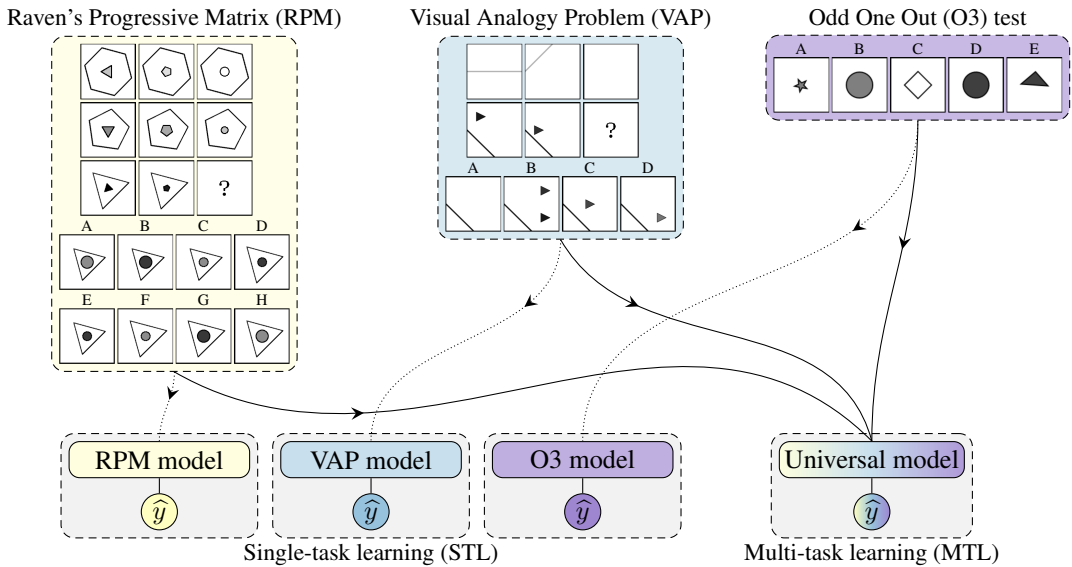
\begin{figure*}[t]
    \centering
    \begin{tikzpicture}
        \begin{scope}[node distance=1pt]
            \node (r1) [panel] {\rpm{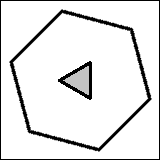}};
            \node (r2) [panel, right=of r1] {\rpm{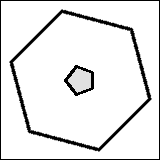}};
            \node (r3) [panel, right=of r2] {\rpm{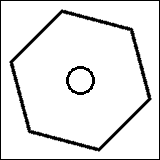}};
            \node (r4) [panel, below=of r1] {\rpm{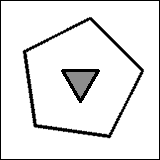}};
            \node (r5) [panel, below=of r2] {\rpm{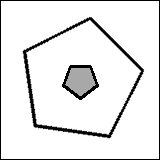}};
            \node (r6) [panel, below=of r3] {\rpm{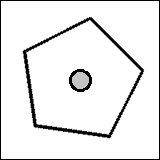}};
            \node (r7) [panel, below=of r4] {\rpm{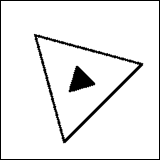}};
            \node (r8) [panel, below=of r5] {\rpm{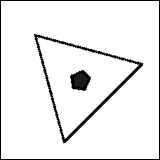}};
            \node (r9) [panel, below=of r6] {\rpm{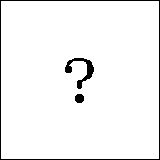}};

            \node (ra1) [panel, below=of r7, xshift=-8pt, yshift=-6pt] {\rpmans{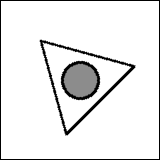}};
            \node (ra2) [panel, right=of ra1] {\rpmans{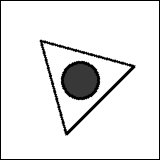}};
            \node (ra3) [panel, right=of ra2] {\rpmans{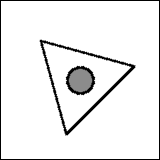}};
            \node (ra4) [panel, right=of ra3] {\rpmans{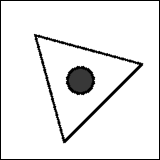}};
            \node (ra5) [panel, below=of ra1, yshift=-6pt] {\rpmans{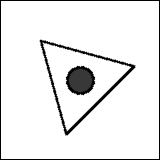}};
            \node (ra6) [panel, right=of ra5] {\rpmans{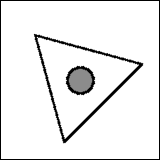}};
            \node (ra7) [panel, right=of ra6] {\rpmans{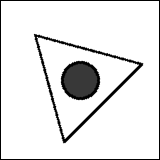}};
            \node (ra8) [panel, right=of ra7] {\rpmans{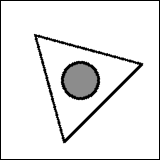}};

            \node (ra1t) [label, above=of ra1] {\tiny A};
            \node (ra2t) [label, above=of ra2] {\tiny B};
            \node (ra3t) [label, above=of ra3] {\tiny C};
            \node (ra4t) [label, above=of ra4] {\tiny D};
            \node (ra5t) [label, above=of ra5] {\tiny E};
            \node (ra6t) [label, above=of ra6] {\tiny F};
            \node (ra7t) [label, above=of ra7] {\tiny G};
            \node (ra8t) [label, above=of ra8] {\tiny H};
        \end{scope}
        \begin{scope}[on background layer]
            \node (rbg) [fit={(r1) (r3) (ra5) (ra8)}, background, fill=rpm!35, inner sep=0.25*\unit] {};
        \end{scope}
        \node [title, anchor=south] at (rbg.north) {Raven's Progressive Matrix (RPM)};

        \begin{scope}[node distance=1pt]
            \node (v1) [panel, right=of r3, xshift=80pt] {\vap{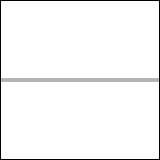}};
            \node (v2) [panel, right=of v1] {\vap{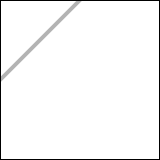}};
            \node (v3) [panel, right=of v2] {\vap{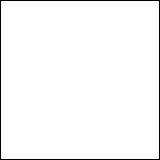}};
            \node (v4) [panel, below=of v1] {\vap{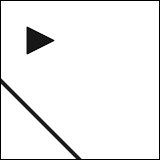}};
            \node (v5) [panel, below=of v2] {\vap{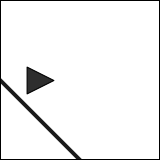}};
            \node (v6) [panel, below=of v3] {\vap{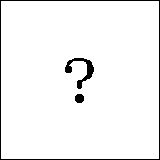}};

            \node (va1) [panel, below=of v4, xshift=-8pt, yshift=-6pt] {\vapans{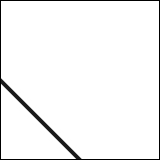}};
            \node (va2) [panel, right=of va1] {\vapans{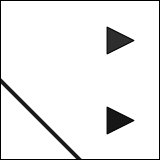}};
            \node (va3) [panel, right=of va2] {\vapans{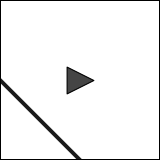}};
            \node (va4) [panel, right=of va3] {\vapans{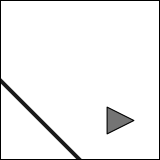}};

            \node (va1t) [label, above=of va1] {\tiny A};
            \node (va2t) [label, above=of va2] {\tiny B};
            \node (va3t) [label, above=of va3] {\tiny C};
            \node (va4t) [label, above=of va4] {\tiny D};
        \end{scope}
        \begin{scope}[on background layer]
            \node (vbg) [fit={(v1) (v3) (va1) (va4)}, background, fill=vap!35, inner sep=0.25*\unit] {};
        \end{scope}
        \node [title, anchor=south] at (vbg.north) {Visual Analogy Problem (VAP)};

        \begin{scope}[node distance=1pt]
            \node (o1) [panel, right=of v3, xshift=50pt, yshift=-5pt] {\ooo{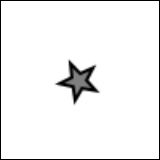}};
            \node (o2) [panel, right=of o1] {\ooo{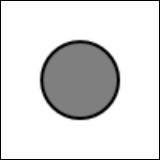}};
            \node (o3) [panel, right=of o2] {\ooo{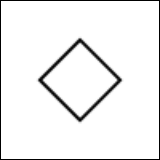}};
            \node (o4) [panel, right=of o3] {\ooo{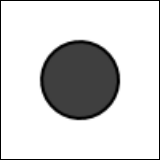}};
            \node (o5) [panel, right=of o4] {\ooo{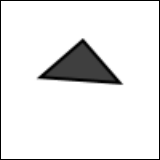}};

            \node (o1t) [label, above=of o1] {\tiny A};
            \node (o2t) [label, above=of o2] {\tiny B};
            \node (o3t) [label, above=of o3] {\tiny C};
            \node (o4t) [label, above=of o4] {\tiny D};
            \node (o5t) [label, above=of o5] {\tiny E};
        \end{scope}
        \begin{scope}[on background layer]
            \node (obg) [fit={(o1) (o1t) (o5) (o5t)}, background, fill=ooo!35, inner sep=0.25*\unit] {};
        \end{scope}
        \node [title, anchor=south] at (obg.north) {\small Odd One Out (O3) test};

        \begin{scope}[node distance=0.5*\unit]
            \node (rm) [layer, fill=rpm!50, below=of rbg, xshift=-0.5*\unit, yshift=-1.8*\unit] {RPM model};
            \node (ry) [embedding, fill=rpm!95, below=of rm] {$\widehat{y}$};

            \node (vm) [layer, fill=vap!50, right=of rm, xshift=0.5*\unit] {VAP model};
            \node (vy) [embedding, fill=vap!95, below=of vm] {$\widehat{y}$};

            \node (om) [layer, fill=ooo!40, right=of vm, xshift=0.5*\unit] {O3 model};
            \node (oy) [embedding, fill=ooo!60, below=of om] {$\widehat{y}$};

            \node (um) [layer, shade, left color=rpm!50, right color=ooo!50, middle color=vap!50, right=of om, xshift=3*\unit] {Universal model};
            \node (uy) [embedding, shade, left color=rpm!95, right color=ooo!60, middle color=vap!95, below=of um] {$\widehat{y}$};
        \end{scope}

        \begin{scope}[on background layer]
            \node (rmbg) [fit={(rm) (ry)}, background, inner sep=0.25*\unit] {};
        \end{scope}
        \begin{scope}[on background layer]
            \node (vmbg) [fit={(vm) (vy)}, background, inner sep=0.25*\unit] {};
        \end{scope}
        \begin{scope}[on background layer]
            \node (ombg) [fit={(om) (oy)}, background, inner sep=0.25*\unit] {};
        \end{scope}
        \node [title, anchor=north] at (vmbg.south) {Single-task learning (STL)};

        \begin{scope}[on background layer]
            \node (umbg) [fit={(um) (uy)}, background, inner sep=0.25*\unit] {};
        \end{scope}
        \node (at) [title, anchor=north] at (umbg.south) {Multi-task learning (MTL)};

        \begin{scope}[decoration={markings, mark=at position 0.4 with {\arrow[scale=1.5,>=stealth]{>}}}] 
            \draw[postaction={decorate}, densely dotted] (rbg.south) to[out=270, in=90] (rm.north);
            \draw[postaction={decorate}, solid] (obg.south) to[out=270, in=90] (um.north);
        \end{scope}
        \begin{scope}[decoration={markings, mark=at position 0.27 with {\arrow[scale=1.5,>=stealth]{>}}}] 
            \draw[postaction={decorate}, densely dotted] (vbg.south) to[out=270, in=90] (vm.north);
            \draw[postaction={decorate}, densely dotted] (obg.south) to[out=270, in=90] (om.north);
            \draw[postaction={decorate}, solid] (rbg.south) to[out=330, in=135] (um.north);
            \draw[postaction={decorate}, solid] (vbg.south) to[out=300, in=105] (um.north);
        \end{scope}
        \draw (rm) -- (ry);
        \draw (vm) -- (vy);
        \draw (om) -- (oy);
        \draw (um) -- (uy);

    \end{tikzpicture}
    \caption{\textbf{STL vs MTL.}
    In this paper, we consider a diverse set of AVR tasks including RPMs, VAPs, and O3 tests (top).
    Existing literature deals with AVR tasks in isolation, which leads to the development of task-specific methods with limited applicability to other, even related problems (bottom left).
    Instead, we propose a single self-configurable model capable of dealing with diverse AVR tasks, thus intrinsically facilitating MTL settings (bottom right).
    }
    \label{fig:intro}
\end{figure*}

Recently, the AVR field has witnessed an increasing interest from the deep learning (DL) community~\cite{malkinski2022review}.
While early works reported successful use of DL models to solve easier instances of AVR tasks, e.g. certain RPMs~\cite{hoshen2017iq,mandziuk2019deepiq}, soon after the effectiveness of DL models in solving more demanding matrices has been questioned~\cite{santoro2018measuring,zhang2019raven}.
Concurrently, a wide suite of AVR benchmarks, extending beyond the concept of RPMs, has been proposed.
For instance, Visual Analogy Problems (VAPs)~\cite{hill2018learning} present a conceptual abstraction challenge similar to RPMs, but with different numbers of context and answer panels.
Another example are Odd One Out (O$3$) tests~\cite{mandziuk2019deepiq} that require the subject to identify the panel that stands out from the remaining ones.
Though all the above AVR problems have abstract reasoning at their core, they differ in the number of panels, task structure, and the rules (visual patterns) that conceptually underlie their matrix representations.

Due to the variety of AVR problems, numerous approaches to tackle them have been proposed~\cite{malkinski2022deep}.
In effect, for the most popular AVR tasks a steady research progress has been observed, and the performance of the methods reached, or even surpassed, the human level~\cite{wu2020scattering}.
Despite impressive performance, contemporary methods largely focus on solving single tasks, often by embedding task-specific design choices into the model's architecture, and cannot be easily evaluated on other, even similar problems.
Consequently, the generality and wider applicability of these solution methods remains unclear.
What is more, the mainstream research commonly utilizes well-established benchmarks with large amounts of available training instances, as the performance of DL AVR methods, when evaluated in low-data regimes, rapidly decreases~\cite{zhuo2021effective}.
On the contrary, human IQ tests which incorporate AVR tasks often assume that the solver has limited prior experience with the tests, enabling less biased evaluation of \emph{fluid intelligence} (as opposed to \emph{crystallised intelligence}, which depends on existing knowledge that can be acquired)~\cite{hofstadter1995fluid}.

To address the above limitations and facilitate development of universal AVR solvers, we:
\begin{enumerate}
    \item Propose a \emph{unified model for solving Single-Choice Abstract visual Reasoning tasks} (SCAR), with no assumptions regarding the number of panels or the task structure and layout.
    The key concept of the model is a linear \textit{Structure-Aware dynamic Layer} (SAL), which adapts its structure to the current problem instance, thus enabling to solve diverse AVR tasks with a common model architecture.
    \item Explore multi-task learning (MTL)~\cite{caruana1997multitask} and transfer learning (TL) formulations in the AVR domain, with knowledge reuse between diverse tasks.
    \item Experimentally evaluate the proposed SCAR and SCL+SAL models on five AVR benchmarks, including RPMs, VAPs, and O$3$ tests, in STL (single-task learning), MTL and TL setups.
    In all three setups, the results prove high effectiveness of SAL-based models, which visibly outperform two universal learning benchmark models, and are on par with the state-of-the-art task-specific SCL model in the case of solving RPMs – the most popular AVR task.
\end{enumerate}

One of the aims of this work is to shift the focus of future AVR research from task-specific methods towards general models applicable to diverse AVR tasks.

\section{Related work}\label{sec:related-work}

\paragraph{Tasks.}
Early AVR works concentrate on RPMs, a well-established benchmark used in measuring human intelligence.
PGM~\cite{santoro2018measuring}, I-RAVEN~\cite{zhang2019raven,hu2021stratified}, and G-set~\cite{mandziuk2019deepiq,tomaszewska2022duel} datasets present RPMs composed of a $3\times3$ grid of context panels, with up to 8 answer panels to choose from.
The VAP benchmark~\cite{hill2018learning}, structurally similar to RPMs, comprises matrices with a $2\times3$ context grid and up to 4 answer panels.
Differently, O$3$ tests~\cite{mandziuk2019deepiq} align up to 6 images in a single row and require the subject to detect the odd panel, which stands out from the remaining ones.
While in this work we concentrate on these five datasets which present sufficient diversity to evaluate knowledge reuse in the AVR domain, there are also other relevant tasks that could be of interest for future work.
In the SVRT dataset~\cite{fleuret2011comparing} context panels are split into two groups, each of them containing shapes with certain attributes that conform to a distinct rule (e.g. one shape inside the other vs two separated shapes).
Given two test panels, the role of the subject is to assign them to a matching group.
A similar setting is considered in Bongard Problems~\cite{bongard1968recognition} and the Bongard-LOGO~\cite{nie2020bongard} dataset.

\paragraph{AVR solvers.}
A frequent design choice made in models for solving AVR tasks is to employ a convolutional backbone which processes each panel separately, producing a vector embedding.
Next, various ways have been explored to aggregate these panel embeddings into the model's prediction.
SRAN~\cite{hu2021stratified} involves three panel representation pathways covering single images, rows/cols, and pairs of rows/cols, each processed with a dedicated convolutional module.
These representations are then aggregated using a sequence of multi-layer perceptrons (MLPs).
SCL~\cite{wu2020scattering} proposes the scattering transformation which merges panel representations with the help of an MLP.
However, together with other related methods~\cite{zheng2019abstract,spratley2020closer,wang2020abstract,benny2021scale,zhuo2021effective}, these approaches incorporate strong task-specific biases into their design (e.g. by utilizing a dense layer with the number of input neurons being determined by the number of panels in the considered problem instance), limiting their applicability to other tasks with a different structure or number of panels.
Some methods take a more universal approach by incorporating neural modules that operate on a set of panel embeddings, without making explicit assumption on the number of input objects.
WReN~\cite{santoro2018measuring} employs the Relation Network (RN)~\cite{santoro2017simple}, which groups input objects into pairs and computes their joint representation.
The Recurrent Neural Network (RNN)~\cite{hill2018learning} aligns inputs in a sequence and processes them successively, updating their internal representation along the way.
Nevertheless, a limitation of these more universal methods is their limited awareness of relative panel positions.
This issue is partly mitigated in~\cite{santoro2018measuring} by concatenating position encodings with panel embeddings.
Contrary to the existing approaches, we propose a model capable of processing AVR tasks with diverse structure and variable number of panels through exploiting a neural layout-aware layer, which dynamically adapts its structure to a particular input instance.

\paragraph{Dynamic neural modules.}
The sample-wise structural layer adaptability of the proposed model relates to the branch of DL research that focuses on designing dynamic neural modules~\cite{han2021dynamic}, which adapt their computation in a sample-dependent manner, as opposed to static modules, which perform the same operation for each data point.
In particular, conditional computation methods adapt the network's architecture by gating access to specific neurons based on the input, leading to reduced inference cost and greater expressivity~\cite{bengio2013estimating,shazeer2017}.
In contrast, the layer proposed in this work in each forward pass utilizes all available parameters in a differentiable manner, alleviating certain optimization challenges related to sparse models~\cite{bengio2015conditional}.
Other dynamic modules adapt kernel sampling locations in convolutional networks~\cite{dai2017deformable,zhu2019deformable}, or directly adjust kernel weights~\cite{gao2020deformable}.
Alternatively, sample-wise direct model parameter prediction has been explored~\cite{schmidhuber1992learning} and has found its applications to linear~\cite{bello2021lambdanetworks} 
 and convolutional~\cite{jia2016dynamic} layers.
In contrast, we rely on a 2D structure of the input to form a linear projection layer with dynamic weights, without taking into account input values, resulting in a simpler and parameter-efficient module.

\section{Method}\label{sec:method}

Considering each AVR dataset as a separate task $t \in \mathcal{T}$, we define it as $t = (\{M_i\}_{i=1}^{N_t}, S_t)$.
Here,
$\{M_i\}_{i=1}^{N_t}$ is a set of $N_t$ problem instances,
$M_i = (X_i, y_i)$ is a specific problem instance,
$X_i = \{x_{i,j}\}_{j=1}^{P_t} = \{x^C_{i,j}\}_{j=1}^{C_t} \cup \{x^A_{i,j}\}_{j=1}^{A_t}$ is a set of $P_t$ matrix panels which can be partitioned into $C_t$ context panels and $A_t$ answer panels,
$x_{i,j} \in [0, 1]^{h \times w}$ is a greyscale image with height $h$ and width $w$,
$y_i \in [1, A_t]$ is an index of the correct answer,
and $S_t$ is the task structure which specifies the layout of the panels (e.g. that context panels are arranged in an $r \times c$ grid, where $r$/$c$ is the number of rows/columns, resp.).
In general, a model for solving matrices from $t$ has the following form:
\begin{equation}
    \mathcal{F}_t (\{\bm{x}_{i,j}\}_{j=1}^{P_t}) = \bm{\widehat{y}}_i
    \label{eq:typical-model-form}
\end{equation}
A common approach to solve single-choice AVR tasks, is to arrange panels $X_i$ into $A_t$ groups $\{ \chi_{i,k} \}_{k=1}^{A_t}$, such that the group $\chi_{i,k} = \{ x_{i,j} \}_{j=1}^{I_t} \subset X_i$ corresponds to the answer $k$, where $I_t$ is the number of panels in the group.
Taking RPMs as an example, for a given RPM instance $M_i$, the groups can be formed by filling in the context grid with each answer panel, giving $\chi_{i,k} = \{ x^C_{i,j} \}_{j=1}^{C_t} \cup \{ x^A_{i,k} \}$ and $I_t = C_t + 1 = 9$.
The group that conforms to the highest number of RPM rules defines the completed form of the (solved) matrix $M_i$.

To reflect this approach in a DL framework, contemporary approaches proposed to:
1)~embed each input panel $x_{i,j}$ with a convolutional encoder $\mathcal{E}_t$ to a latent representation $h_{i,j}\in\mathbb{R}^{d_h}$,
2)~arrange the embeddings into $A_t$ groups $\{ H_{i,k} \}_{k=1}^{A_t}$, where group $H_{i,k} \subset \{ h_{i,j} \}_{j=1}^{P_t}$ corresponds to answer $k$,
3)~fuse the embeddings inside each group with a reasoning module $\mathcal{G}_t$ to generate a joint embedding $g_{i,k} \in \mathbb{R}^{d_g}$,
4)~employ a decoder $\mathcal{D}$ to convert the joint embedding to an alignment score $s_{i,k} \in \mathbb{R}$,
5)~transform the resultant scores into a probability distribution $\widehat{p}_i$ over the set of available answers,
6)~return the answer corresponding to the highest probability as the model's prediction $\widehat{y}_i$:
\begin{align}
    &1)\ \mathcal{E}_t (\bm{x}_{i,j}) = \bm{h}_{i,j}
    \ 
    2)\ \textsc{Arrange} ( \{ \bm{h}_{i,j} \}_{j=1}^{P_t} ) = \{ \bm{H}_{i,k} \}_{k=1}^{A_t}
    \nonumber
    \\
    &3)\ \mathcal{G}_t ( \bm{H}_{i,k} \ \vert \ S_t) = \bm{g}_{i,k}
    \label{eq:typical-model-components}
    \quad
    4)\ \mathcal{D}_t ( \bm{g}_{i,k} ) = \bm{s}_{i,k}
    \\
    &5)\ \textsc{softmax} ( \{ \bm{s}_{i,k} \}_{k=1}^{A_t}) = \bm{\widehat{p}}_i
    \nonumber
    \quad
    6)\ \textsc{arg}\,\textsc{max}\ \bm{\widehat{p}}_i = \bm{\widehat{y}}_i
\end{align}
Insofar, models were developed with the aim of tackling a specific task, without much consideration for generalizing to other problems.
To a large extent, this led to task-specific models which assume $I_t$ and $S_t$ in advance, and incorporate them directly into their architecture.
To lift these assumptions and bolster universality of AVR models, we
i) propose a neural module which dynamically adapts to matrices with various $I_t$ and $S_t$,
ii) embody it in an end-to-end architecture for solving AVR tasks, and
iii) demonstrate how the method can be applied to solving tasks with diverse structures.

\definecolor{rpm}{HTML}{FFFFBF}
\definecolor{vap}{HTML}{91BFDB}

\tikzstyle{background} = [draw, fill=black!5, rounded corners=5pt, densely dashed, inner sep=0]
\tikzstyle{panel} = [thick, rectangle, draw=black, inner sep=0]
\tikzstyle{label} = [rectangle, inner sep=0]
\tikzstyle{dim} = [inner sep=1, align=center, font={\fontsize{4}{5}\selectfont}]

\tikzstyle{embedding} = [circle, text centered, draw=black, fill=layer!50, inner sep=0, minimum height=1.2*\unit, minimum width=1.2*\unit]

\tikzstyle{arrow} = [->,>={Latex[scale=1]},rounded corners=5pt]

\newcommand{\drawbox}[5][]{
    \pgfmathsetmacro \a {0.3} 
    \pgfmathsetmacro \angle {30}
    \pgfmathsetmacro \xd {{2/3*cos(\angle)*\a*3}}
    \pgfmathsetmacro \yd {{2/3*sin(\angle)*\a*3}}
    \pgfmathsetmacro \x {{#2*\a-\a-(#3*\a-\a)*(\xd)}}
    \pgfmathsetmacro \y {{#4*\a-\a+(#3*\a-\a)*(\yd)}}

    \draw[fill=#5, very thin] (\x,\y) -- (\x+\a,\y) -- (\x+\a,\y+\a) -- (\x,\y+\a) -- cycle; 
    \draw[fill=#5, very thin] (\x,\y+\a) -- (\x-\xd,\y+\a+\yd) -- (\x+\a-\xd,\y+\a+\yd) -- (\x+\a,\y+\a) -- cycle; 
    \draw[fill=#5, very thin] (\x,\y+\a) -- (\x-\xd,\y+\a+\yd) -- (\x-\xd,\y+\yd) -- (\x,\y) -- cycle; 
    \node[inner sep=0, anchor=center] at (\x+\a*0.5, \y+\a*0.5) {\tiny#1};
}

\newcommand{\rect}[4][]{
    \pgfmathsetmacro \a {0.2} 
    \pgfmathsetmacro \x {{#2*\a-\a}}
    \pgfmathsetmacro \y {{#3*\a-\a}}

    \draw[fill=#4, very thin] (\x,\y) -- (\x+\a,\y) -- (\x+\a,\y+\a) -- (\x,\y+\a) -- cycle;
    \node[inner sep=0, anchor=center] at (\x+\a*0.5, \y+\a*0.5) {\tiny#1};
}

\newcommand{\rec}[4][]{
    \pgfmathsetmacro \a {0.2} 
    \pgfmathsetmacro \x {{#3*\a-\a}}
    \pgfmathsetmacro \y {{-#2*\a+\a}}

    \draw[fill=#4, very thin] (\x,\y) -- (\x+\a,\y) -- (\x+\a,\y-\a) -- (\x,\y-\a) -- cycle;
    \node[inner sep=0, anchor=center] at (\x+\a*0.5, \y-\a*0.5) {\tiny#1};
}

\newcommand{\brect}[4]{
    \pgfmathsetmacro \a {0.2} 
    \pgfmathsetmacro \dy {\a*#3}
    \pgfmathsetmacro \dx {\a*#4}
    \pgfmathsetmacro \x {{#2*\dx-\dx}}
    \pgfmathsetmacro \y {{-#1*\dy+\dy}}

    \draw[thick] (\x,\y) -- (\x+\dx,\y) -- (\x+\dx,\y-\dy) -- (\x,\y-\dy) -- cycle;
}

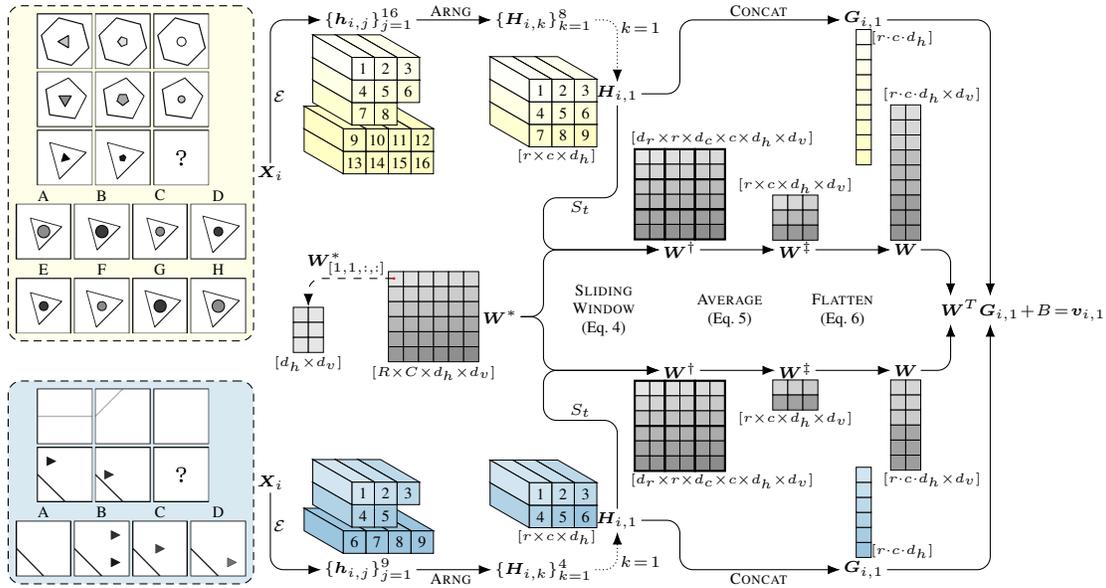
\begin{figure*}[t]
    \centering
    \begin{tikzpicture}
        \begin{scope}[node distance=1pt]
            \node (r1) [panel] {\rpm{0}};
            \node (r2) [panel, right=of r1] {\rpm{1}};
            \node (r3) [panel, right=of r2] {\rpm{2}};
            \node (r4) [panel, below=of r1] {\rpm{3}};
            \node (r5) [panel, below=of r2] {\rpm{4}};
            \node (r6) [panel, below=of r3] {\rpm{5}};
            \node (r7) [panel, below=of r4] {\rpm{6}};
            \node (r8) [panel, below=of r5] {\rpm{7}};
            \node (r9) [panel, below=of r6] {\rpm{8}};

            \node (ra1) [panel, below=of r7, xshift=-8pt, yshift=-6pt] {\rpmans{0}};
            \node (ra2) [panel, right=of ra1] {\rpmans{1}};
            \node (ra3) [panel, right=of ra2] {\rpmans{2}};
            \node (ra4) [panel, right=of ra3] {\rpmans{3}};
            \node (ra5) [panel, below=of ra1, yshift=-6pt] {\rpmans{4}};
            \node (ra6) [panel, right=of ra5] {\rpmans{5}};
            \node (ra7) [panel, right=of ra6] {\rpmans{6}};
            \node (ra8) [panel, right=of ra7] {\rpmans{7}};

            \node (ra1t) [label, above=of ra1] {\tiny A};
            \node (ra2t) [label, above=of ra2] {\tiny B};
            \node (ra3t) [label, above=of ra3] {\tiny C};
            \node (ra4t) [label, above=of ra4] {\tiny D};
            \node (ra5t) [label, above=of ra5] {\tiny E};
            \node (ra6t) [label, above=of ra6] {\tiny F};
            \node (ra7t) [label, above=of ra7] {\tiny G};
            \node (ra8t) [label, above=of ra8] {\tiny H};
        \end{scope}
        \begin{scope}[on background layer]
            \node (rbg) [fit={(r1) (r3) (ra5) (ra8)}, background, fill=rpm!35, inner sep=0.25*\unit] {};
        \end{scope}

        \begin{scope}[node distance=1pt]
            \node (v1) [panel, below=of r1, yshift=-120pt] {\vap{0}};
            \node (v2) [panel, right=of v1] {\vap{1}};
            \node (v3) [panel, right=of v2] {\vap{2}};
            \node (v4) [panel, below=of v1] {\vap{3}};
            \node (v5) [panel, below=of v2] {\vap{4}};
            \node (v6) [panel, below=of v3] {\vap{5}};

            \node (va1) [panel, below=of v4, xshift=-8pt, yshift=-6pt] {\vapans{0}};
            \node (va2) [panel, right=of va1] {\vapans{1}};
            \node (va3) [panel, right=of va2] {\vapans{2}};
            \node (va4) [panel, right=of va3] {\vapans{3}};

            \node (va1t) [label, above=of va1] {\tiny A};
            \node (va2t) [label, above=of va2] {\tiny B};
            \node (va3t) [label, above=of va3] {\tiny C};
            \node (va4t) [label, above=of va4] {\tiny D};
        \end{scope}
        \begin{scope}[on background layer]
            \node (vbg) [fit={(v1) (v3) (va1) (va4)}, background, fill=vap!35, inner sep=0.25*\unit] {};
        \end{scope}

        \node [anchor=north, xshift=90pt, inner sep=1pt] (rpel) at (rbg.north) {\tiny $\{\bm{h}_{i,j}\}_{j=1}^{16}$};
        \node [anchor=west, inner sep=1pt] (rl) at (rbg.east) {\tiny $\bm{X}_i$};
        \node [below=of rpel, yshift=5pt] (rae) {
            \begin{tikzpicture}
                \drawbox[16]{4}{1}{1}{rpm!90}
                \drawbox[15]{3}{1}{1}{rpm!90}
                \drawbox[14]{2}{1}{1}{rpm!90}
                \drawbox[13]{1}{1}{1}{rpm!90}
                \drawbox[12]{4}{1}{2}{rpm!90}
                \drawbox[11]{3}{1}{2}{rpm!90}
                \drawbox[10]{2}{1}{2}{rpm!90}
                \drawbox[9]{1}{1}{2}{rpm!90}
            \end{tikzpicture}
        };
        \node [below=of rpel, xshift=-1pt, yshift=32pt] (rpe) {
            \begin{tikzpicture}
                \drawbox[8]{2}{1}{1}{rpm!80}
                \drawbox[7]{1}{1}{1}{rpm!70}
                \drawbox[6]{3}{1}{2}{rpm!60}
                \drawbox[5]{2}{1}{2}{rpm!50}
                \drawbox[4]{1}{1}{2}{rpm!40}
                \drawbox[3]{3}{1}{3}{rpm!30}
                \drawbox[2]{2}{1}{3}{rpm!20}
                \drawbox[1]{1}{1}{3}{rpm!10}
            \end{tikzpicture}
        };

        \node [right=of rpel, inner sep=1pt] (rgel) {\tiny $\{\bm{H}_{i,k}\}_{k=1}^{8}$};
        \node [anchor=north, yshift=-5pt] (rge1) at (rgel.south) {
            \begin{tikzpicture}
                \drawbox[9]{3}{1}{1}{rpm!90}
                \drawbox[8]{2}{1}{1}{rpm!80}
                \drawbox[7]{1}{1}{1}{rpm!70}
                \drawbox[6]{3}{1}{2}{rpm!60}
                \drawbox[5]{2}{1}{2}{rpm!50}
                \drawbox[4]{1}{1}{2}{rpm!40}
                \drawbox[3]{3}{1}{3}{rpm!30}
                \drawbox[2]{2}{1}{3}{rpm!20}
                \drawbox[1]{1}{1}{3}{rpm!10}
            \end{tikzpicture}
        };
        \node [anchor=west, inner sep=1pt, xshift=-5pt, yshift=3pt] (rgel1) at (rge1.east) {\tiny $\bm{H}_{i,1}$};
        \node [dim, anchor=north, xshift=5pt, yshift=4pt] at (rge1.south) {$[r{\times}c{\times}d_h]$};

        \node [anchor=south, inner sep=1pt] at ($ (rpel |- vbg.south) $) (vpel) {\tiny $\{\bm{h}_{i,j}\}_{j=1}^{9}$};
        \node [anchor=west, inner sep=1pt] (vl) at (vbg.east) {\tiny $\bm{X}_i$};
        \node [anchor=south, yshift=-3pt] (vae) at (vpel.north) {
            \begin{tikzpicture}
                \drawbox[9]{4}{1}{1}{vap!90}
                \drawbox[8]{3}{1}{1}{vap!90}
                \drawbox[7]{2}{1}{1}{vap!90}
                \drawbox[6]{1}{1}{1}{vap!90}
            \end{tikzpicture}
        };
        \node [xshift=-1pt, yshift=2pt] (vpe) at (vae.north) {
            \begin{tikzpicture}
                \drawbox[5]{2}{1}{1}{vap!80}
                \drawbox[4]{1}{1}{1}{vap!70}
                \drawbox[3]{3}{1}{2}{vap!60}
                \drawbox[2]{2}{1}{2}{vap!50}
                \drawbox[1]{1}{1}{2}{vap!40}
            \end{tikzpicture}
        };

        \node [right=of vpel, inner sep=1pt] (vgel) {\tiny $\{\bm{H}_{i,k}\}_{k=1}^{4}$};
        \node [anchor=south, yshift=7pt] (vge1) at (vgel.north) {
            \begin{tikzpicture}
                \drawbox[6]{3}{1}{1}{vap!90}
                \drawbox[5]{2}{1}{1}{vap!80}
                \drawbox[4]{1}{1}{1}{vap!70}
                \drawbox[3]{3}{1}{2}{vap!60}
                \drawbox[2]{2}{1}{2}{vap!50}
                \drawbox[1]{1}{1}{2}{vap!40}
            \end{tikzpicture}
        };
        \node [anchor=west, inner sep=1pt, xshift=-5pt, yshift=-10pt] (vgel1) at (vge1.east) {\tiny $\bm{H}_{i,1}$};
        \node [dim, anchor=north, xshift=5pt, yshift=4pt] at (vge1.south) {$[r{\times}c{\times}d_h]$};

        \node [inner sep=0, xshift=25pt] (wm) at ($ (rae.south)!0.5!(vpe.north) $) {
            \begin{tikzpicture}
                \foreach \i in {1,...,6} {
                    \foreach \j in {1,...,6} {
                        \pgfmathsetmacro{\c}{20+(\i-1)*12+(\j-1)*2};
                        \rec[]{\i}{\j}{gray!\c};
                    }
                }
            \end{tikzpicture}
        };
        \node [dim, anchor=north] at (wm.south) {$[R{\times}C{\times}d_h{\times}d_v]$};
        \node [inner sep=1pt, anchor=west] (wml) at (wm.east) {\tiny $\bm{W}^*$};
        \node (wn1) at ($ (rgel |- wml) $) {};
        \node [yshift=-5pt] (swr) at ($ (wn1)!0.5!(rge1.south) $) {};
        \node [yshift=5pt] (swv) at ($ (wn1)!0.5!(vge1.north) $) {};

        \node [inner sep=0, anchor=east, xshift=-24pt, yshift=-5pt] (wmc) at (wm.west) {
            \begin{tikzpicture}
                \foreach \i in {1,...,3} {
                    \foreach \j in {1,...,2} {
                        \rec[]{\i}{\j}{gray!20};
                    }
                }
            \end{tikzpicture}
        };
        \node [dim, anchor=north] at (wmc.south) {$[d_h{\times}d_v]$};
        \node (wmcell1) at ($ (wm.north west)!1/6!(wm.west) $) {};
        \node (wmcell2) at ($ (wm.north west)!1/6!(wm.north) $) {};
        \node (wmcell) at ($ (wmcell2 |- wmcell1) $) {};
        \node (wmcell3) at ($ (wmc |- wmcell) $) {};
        \node [inner sep=1, align=center, anchor=south, xshift=-2pt] at ($ (wmcell3)!0.5!(wmcell) $) {\tiny$\bm{W}^*_{[1, 1, :, :]}$};
        \draw [arrow, dashed, {Circle[red,length=1pt]}-Latex] (wmcell.center) -- (wmcell3.center) -- (wmc);

        \node [inner sep=1pt, right=of swr, xshift=12pt] (wmrl) {\tiny$\bm{W}^{\dagger}$};
        \node [inner sep=1pt, right=of swv, xshift=12pt] (wmvl) {\tiny$\bm{W}^{\dagger}$};
        \node (swrc) at ($ (swr)!0.5!(wmrl) $) {};
        \node (swvc) at ($ (swv)!0.5!(wmvl) $) {};
        \node [inner sep=1pt, font={\tiny}, align=center, xshift=-3pt] at ($ (swrc)!0.5!(swvc) $) {\sc{Sliding}\\\sc{Window}\\(Eq.~\ref{eq:sliding-window})};
        \node [anchor=south] at (wmrl) (wmr) {
            \begin{tikzpicture}
                \foreach \i in {1,...,6} {
                    \foreach \j in {1,...,6} {
                        \pgfmathsetmacro{\c}{20+(\i-1)*12+(\j-1)*2};
                        \rec[]{\i}{\j}{gray!\c};
                    }
                }
                \foreach \i in {1,...,3} {
                    \foreach \j in {1,...,3} {
                        \brect{\i}{\j}{2}{2};
                    }
                }
            \end{tikzpicture}
        };
        \node [dim, anchor=south, xshift=16pt, yshift=-4pt] at (wmr.north) {$[d_r{\times}r{\times}d_c{\times}c{\times}d_h{\times}d_v]$};
        \node [anchor=north] at (wmvl) (wmv) {
            \begin{tikzpicture}
                \foreach \i in {1,...,6} {
                    \foreach \j in {1,...,6} {
                        \pgfmathsetmacro{\c}{20+(\i-1)*12+(\j-1)*2};
                        \rec[]{\i}{\j}{gray!\c};
                    }
                }
                \foreach \i in {1,...,2} {
                    \foreach \j in {1,...,3} {
                        \brect{\i}{\j}{3}{2};
                    }
                }
            \end{tikzpicture}
        };
        \node [dim, anchor=north, xshift=16pt, yshift=4pt] at (wmv.south) {$[d_r{\times}r{\times}d_c{\times}c{\times}d_h{\times}d_v]$};

        \node [inner sep=1pt, right=of wmrl] (wmarl) {\tiny$\bm{W}^{\ddagger}$};
        \node [inner sep=1pt, right=of wmvl] (wmavl) {\tiny$\bm{W}^{\ddagger}$};
        \node (wmarc) at ($ (wmrl)!0.5!(wmarl) $) {};
        \node (wmavc) at ($ (wmvl)!0.5!(wmavl) $) {};
        \node [inner sep=1pt, font={\tiny}, align=center, xshift=-3pt] at ($ (wmarc)!0.5!(wmavc) $) {\sc{Average}\\(Eq.~\ref{eq:average})};
        \node [anchor=south] (wmar) at (wmarl) {
            \begin{tikzpicture}
                \foreach \i in {1,...,3} {
                    \foreach \j in {1,...,3} {
                        \pgfmathsetmacro{\c}{27+(\i-1)*24+(\j-1)*4};
                        \rec[]{\i}{\j}{gray!\c};
                    }
                }
            \end{tikzpicture}
        };
        \node [dim, anchor=south, yshift=-4pt] at (wmar.north) {$[r{\times}c{\times}d_h{\times}d_v]$};
        \node [anchor=north] (wmav) at (wmavl) {
            \begin{tikzpicture}
                \foreach \i in {1,...,2} {
                    \foreach \j in {1,...,3} {
                        \pgfmathsetmacro{\c}{33+(\i-1)*36+(\j-1)*4};
                        \rec[]{\i}{\j}{gray!\c};
                    }
                }
            \end{tikzpicture}
        };
        \node [dim, anchor=north, yshift=4pt] at (wmav.south) {$[r{\times}c{\times}d_h{\times}d_v]$};

        \node [inner sep=1pt, right=of wmarl] (wmfrl) {\tiny$\bm{W}$};
        \node [inner sep=1pt, right=of wmavl] (wmfvl) {\tiny$\bm{W}$};
        \node (wmfrc) at ($ (wmarl)!0.5!(wmfrl) $) {};
        \node (wmfvc) at ($ (wmavl)!0.5!(wmfvl) $) {};
        \node (wmfcl) [inner sep=1pt, font={\tiny}, align=center, xshift=-3pt] at ($ (wmfrc)!0.5!(wmfvc) $) {\sc{Flatten}\\(Eq.~\ref{eq:flatten})};
        \node [anchor=south] (wmfr) at (wmfrl) {
            \begin{tikzpicture}
                \foreach \i in {1,...,3} {
                    \foreach \j in {1,...,3} {
                        \pgfmathsetmacro{\c}{27+(\i-1)*24+(\j-1)*4};
                        \pgfmathsetmacro{\row}{(\i-1)*3+\j};
                        \rec[]{\row}{1}{gray!\c};
                        \rec[]{\row}{2}{gray!\c};
                    }
                }
            \end{tikzpicture}
        };
        \node [dim, anchor=south, xshift=10pt, yshift=-4pt] at (wmfr.north) {$[r{\cdot}c{\cdot}d_h{\times}d_v]$};
        \node [anchor=north] (wmfv) at (wmfvl) {
            \begin{tikzpicture}
                \foreach \i in {1,...,2} {
                    \foreach \j in {1,...,3} {
                        \pgfmathsetmacro{\c}{33+(\i-1)*36+(\j-1)*4};
                        \pgfmathsetmacro{\row}{(\i-1)*3+\j};
                        \rec[]{\row}{1}{gray!\c};
                        \rec[]{\row}{2}{gray!\c};
                    }
                }
            \end{tikzpicture}
        };
        \node [dim, anchor=north, xshift=10pt, yshift=4pt] at (wmfv.south) {$[r{\cdot}c{\cdot}d_h{\times}d_v]$};

        \node [inner sep=1pt, xshift=8pt] (rgecl) at ($ (wmfcl |- rgel) $) {\tiny$\bm{G}_{i,1}$};
        \node (rgecl1) at ($ (wmrl |- rgel1) $) {};
        \node (rgecl2) at ($ (wmrl |- rgel) $) {};
        \node [inner sep=1pt, xshift=8pt] (vgecl) at ($ (wmfcl |- vgel) $) {\tiny$\bm{G}_{i,1}$};
        \node (vgecl1) at ($ (wmvl |- vgel1) $) {};
        \node (vgecl2) at ($ (wmvl |- vgel) $) {};
        \node [anchor=north] (rgec) at (rgecl) {
            \begin{tikzpicture}
                \rect[]{1}{1}{rpm!90};
                \rect[]{1}{2}{rpm!80};
                \rect[]{1}{3}{rpm!70};
                \rect[]{1}{4}{rpm!60};
                \rect[]{1}{5}{rpm!50};
                \rect[]{1}{6}{rpm!40};
                \rect[]{1}{7}{rpm!30};
                \rect[]{1}{8}{rpm!20};
                \rect[]{1}{9}{rpm!10};
            \end{tikzpicture}
        };
        \node [dim, anchor=west, xshift=-5pt, yshift=-7pt] at (rgec.north east) {$[r{\cdot}c{\cdot}d_h]$};
        \node [anchor=south, yshift=1pt] (vgec) at (vgecl) {
            \begin{tikzpicture}
                \rect[]{1}{1}{vap!90};
                \rect[]{1}{2}{vap!80};
                \rect[]{1}{3}{vap!70};
                \rect[]{1}{4}{vap!60};
                \rect[]{1}{5}{vap!50};
                \rect[]{1}{6}{vap!40};
            \end{tikzpicture}
        };
        \node [dim, anchor=west, xshift=-5pt, yshift=6pt] at (vgec.south east) {$[r{\cdot}c{\cdot}d_h]$};

        \node (dpc) at ($ (wmfrl)!0.5!(wmfvl) $) {};
        \node [inner sep=1pt, right=of dpc, xshift=-20pt] (dpl) {\tiny$\bm{W}^T\bm{G}_{i,1}\!+\!B\!=\!\bm{v}_{i,1}$};
        \node [xshift=5pt] (dpr) at ($ (dpl.west |- wmfrl) $) {};
        \node [xshift=5pt] (dpv) at ($ (dpl.west |- wmfvl) $) {};
        \node [xshift=20pt] (dpgr) at ($ (dpl.west |- rgecl) $) {};
        \node [xshift=20pt] (dpgv) at ($ (dpl.west |- vgecl) $) {};

        \node (rlc) at ($ (rl |- rpel) $) {};
        \node (rgc) at ($ (rgel1 |- rgel) $) {};
        \node [inner sep=1pt, anchor=west, align=center, yshift=-3pt] at (rgc) {\tiny$k\!=\!1$};
        \node (rgswc) at ($ (swr)!1/3!(rgel1) $) {};
        \node (rgswc1) at ($ (rgel1 |- rgswc) $) {};
        \node (rgswc2) at ($ (swr |- rgswc) $) {};
        \draw [arrow] (rl) -- (rlc.center) -- (rpel);
        \draw [arrow] (rpel) -- (rgel);
        \draw [arrow, densely dotted] (rgel) -- (rgc.center) -- (rgel1);
        \draw [arrow] (rgel1) -- (rgswc1.center) -- (rgswc2.center) -- (swr.center) -- (wmrl);
        \draw [arrow] (rgel1) -- (rgecl1.center) -- (rgecl2.center) -- (rgecl);
        \draw [arrow] (wmrl) -- (wmarl);
        \draw [arrow] (wmarl) -- (wmfrl);
        \draw [arrow] (wmfrl) -- (dpr.center) -- ($ (dpr |- dpl.north) $);
        \draw [arrow] (rgecl) -- (dpgr.center) -- ($ (dpgr |- dpl.north) $);
        \node [inner sep=1pt, anchor=west, font={\tiny}, align=center] at ($ (rl)!0.5!(rlc) $) {$\mathcal{E}$};
        \node [inner sep=1pt, anchor=south, font={\tiny}, align=center, xshift=-2pt] at ($ (rpel)!0.5!(rgel) $) {\sc{Arng}};
        \node [inner sep=1pt, anchor=south, font={\tiny}, align=center, xshift=-5pt] at ($ (rgecl2)!0.5!(rgecl) $) {\sc{Concat}};
        \node [inner sep=1pt, anchor=north, font={\tiny}, align=center] at ($ (rgswc1)!0.5!(rgswc2) $) {$S_t$};

        \node (vlc) at ($ (vl |- vpel) $) {};
        \node (vgc) at ($ (vgel1 |- vgel) $) {};
        \node [inner sep=1pt, anchor=west, align=center, yshift=3pt] at (vgc) {\tiny$k\!=\!1$};
        \node (vgswc) at ($ (swv)!1/3!(vgel1) $) {};
        \node (vgswc1) at ($ (vgel1 |- vgswc) $) {};
        \node (vgswc2) at ($ (swv |- vgswc) $) {};
        \draw [arrow] (vl) -- (vlc.center) -- (vpel);
        \draw [arrow] (vpel) -- (vgel);
        \draw [arrow, densely dotted] (vgel) -- (vgc.center) -- (vgel1);
        \draw [arrow] (vgel1) -- (vgswc1.center) -- (vgswc2.center) -- (swv.center) -- (wmvl);
        \draw [arrow] (vgel1) -- (vgecl1.center) -- (vgecl2.center) -- (vgecl);
        \draw [arrow] (wmvl) -- (wmavl);
        \draw [arrow] (wmavl) -- (wmfvl);
        \draw [arrow] (wmfvl) -- (dpv.center) -- ($ (dpv |- dpl.south) $);
        \draw [arrow] (vgecl) -- (dpgv.center) -- ($ (dpgv |- dpl.south) $);
        \node [inner sep=1pt, anchor=west, font={\tiny}, align=center] at ($ (vl)!0.5!(vlc) $) {$\mathcal{E}$};
        \node [inner sep=1pt, anchor=north, font={\tiny}, align=center, xshift=-2pt] at ($ (vpel)!0.5!(vgel) $) {\sc{Arng}};
        \node [inner sep=1pt, anchor=north, font={\tiny}, align=center, xshift=-5pt] at ($ (vgecl2)!0.5!(vgecl) $) {\sc{Concat}};
        \node [inner sep=1pt, anchor=south, font={\tiny}, align=center] at ($ (vgswc1)!0.5!(vgswc2) $) {$S_t$};

        \draw [arrow] (wml.east) -- (wn1.center) -- (swr.center) -- (wmrl);
        \draw [arrow] (wml.east) -- (wn1.center) -- (swv.center) -- (wmvl);

    \end{tikzpicture}
    \caption{
    \textbf{Structure-Aware dynamic Layer.}
    SAL enables processing of AVR tasks with diverse structures by adapting its weights to the problem instance.
    The figure illustrates two separate forward passes through the layer for an RPM (top) and a VAP matrix (bottom).
    From left, the panels $X_i$ belonging to a single problem instance are processed with an encoder $\mathcal{E}$.
    The resultant embeddings $\{h_{i,j}\}$ are arranged into $A_t$ groups $\{H_{i,k}\}$.
    Next, the figure illustrates further processing of a single group $H_{i,1}$ for both matrices.
    Thanks to SAL's adaptability, the resultant vector $v_{i,1} \in \mathbb{R}^{d_v}$ has the same dimensions irrespectively of the considered task, thus enabling uniform processing in the subsequent model layers.
    }
    \label{fig:layer}
\end{figure*}

\subsection{Dynamic weight adaptation}
To solve various AVR tasks with a single unified model architecture, we start by designing a differentiable computation layer $\psi ( H_{i,k}\ \vert\ S_t ) = v_{i,k}$ which produces a latent representation $v_{i,k} \in \mathbb{R}^{d_v}$ for a set of $I_t$ panel embeddings $H_{i,k}$, with a layout defined by the task structure $S_t$.
As we discuss in Section~\ref{subsec:model}, $\psi$ will serve as the first layer in $\mathcal{G}$, and $v_{i,k}$ will be transformed to $g_{i,k}$ by the subsequent layers of $\mathcal{G}$.

To instantiate $\psi$, we propose the \emph{Structure-Aware dynamic Layer} (SAL) -- see Fig.~\ref{fig:layer}:
\begin{align}
    \mathcal{W} (W^*\ |\ S_t) &= W
    \qquad
    \textsc{Concat}( \bm{H}_{i,k} ) = \bm{G}_{i,k}
    \nonumber
    \\
    W^T \bm{G}_{i,k} + B &= \bm{v}_{i,k}
    \label{eq:dynamic-weight-adaptation-linear}
\end{align}
From a high-level perspective, SAL can be viewed as a linear layer with weights  $W \in \mathbb{R}^{r\cdot c\cdot d_h \times d_v}$, where $r$/$c$ is the number of rows/columns in the context grid, and $B \in \mathbb{R}^{d_v}$ are optional biases.
In contrast to the widely-adopted linear layer with static weights, the matrix $W$ in SAL is computed dynamically by the transformation $\mathcal{W}$, based on the underlying weight matrix $W^* \in \mathbb{R}^{R \times C \times d_h \times d_v}$, where $R$ and $C$ are its row and column dimensions, resp.
Specifically, $\mathcal{W}$ utilizes a sliding window, which is adapted according to the task structure $S_t$, to unfold the first two dimensions of $W$ from $R \times C$ to $d_r \times r \times d_c \times c$ (Eq.~\ref{eq:sliding-window}), averages along the first and third dimensions (Eq.~\ref{eq:average}), and flattens the matrix into 2D form (Eq.~\ref{eq:flatten}):
\begin{align}
    \textsc{SlidingWindow} (W^* \vert S_t) &= W^\dagger \in \mathbb{R}^{d_r \times r \times d_c \times c \times d_h \times d_v}
    \label{eq:sliding-window}
    \\
    \frac{1}{d_r} \frac{1}{d_c} \sum^{d_r}_{i=1} \sum^{d_c}_{k=1} W^\dagger_{ijklmn} &= W^\ddagger_{jlmn} \in \mathbb{R}^{r \times c \times d_h \times d_v}
    \label{eq:average}
    \\
    \textsc{Flatten} ( W^\ddagger, 1:3 ) &= W \in \mathbb{R}^{r\cdot c\cdot d_h \times d_v}
    \label{eq:flatten}
\end{align}
where $d_r = \frac{R}{r}$ and $d_c = \frac{C}{c}$.
The SAL input is constructed by concatenating the embedding vectors $H_{i,k}$, to obtain $G_{i,k} \in \mathbb{R}^{r\cdot c\cdot d_h}$.

\paragraph{Multi-head SAL.}
On top of the default formulation, we propose a multi-head version of SAL.
In this setting, each embedding vector $h_{i,j}$ from $\{h_{i,j}\}_{j=1}^{P_t}$ is partitioned into a set of $L$ vectors $\{h_{i,j,l}\}_{l=1}^L$, where $h_{i,j,l} = h_{i,j}[(l-1)d_l:ld_l] \in \mathbb{R}^{d_l}$, and $d_l = \frac{d_h}{L}$.
The resultant set of panel embeddings $\{\{h_{i,j,l}\}_{l=1}^L\}_{j=1}^{P_t}$ is processed in parallel for each $l$ with the arrangement operator (\textsc{Arrange} in Eq.~\ref{eq:typical-model-components}) and $\psi$, leading to $\{v_{i,k,l}\}_{l=1}^L$.
The vectors are then concatenated producing $v_{i,k}$, which can be processed by the subsequent layers of $\mathcal{G}$.
The above multi-head SAL variant, with $L=20$, was used in all conducted experiments.

\definecolor{layer}{HTML}{E1E1E1}
\definecolor{activation}{HTML}{A9C8C0}
\definecolor{conv}{HTML}{DBBC8E}
\definecolor{linear}{HTML}{D9C2BD}
\definecolor{layernorm}{HTML}{B8E0F6}
\definecolor{batchnorm}{HTML}{A4CCE3}
\definecolor{feedforward}{HTML}{91BFDB}
\definecolor{dfp}{HTML}{E6A57E}

\newcommand{\panel}[1]{\includegraphics[width=0.04\textwidth]{images/rpm/context_#1}}

\tikzstyle{layer} = [rectangle, rounded corners, text centered, draw=black, fill=layer, inner sep=0, minimum width=5*\unit, minimum height=1.2*\unit]
\tikzstyle{label} = [align=center, font={\footnotesize}]
\tikzstyle{activation} = [layer, fill=activation!50]
\tikzstyle{conv2d} = [layer, fill=conv!50]
\tikzstyle{conv1d} = [layer, fill=conv!35]
\tikzstyle{linear} = [layer, fill=linear!50]
\tikzstyle{layernorm} = [layer, fill=layernorm!35]
\tikzstyle{batchnorm} = [layer, fill=batchnorm!50]
\tikzstyle{feedforward} = [layer, fill=linear]
\tikzstyle{emptylayer} = [rectangle, inner sep=0, minimum width=5*\unit, minimum height=1.2*\unit]
\tikzstyle{background} = [draw, fill=black!5, rounded corners=5pt, densely dashed, inner sep=0]

\tikzstyle{embedding} = [rectangle, rounded corners=6pt, text centered, draw=black, fill=layer!50, inner sep=0, minimum width=2*\unit, minimum height=1.2*\unit]
\tikzstyle{panel} = [thick, rectangle, draw=black, inner sep=0]

\tikzstyle{arrow} = [-Stealth]

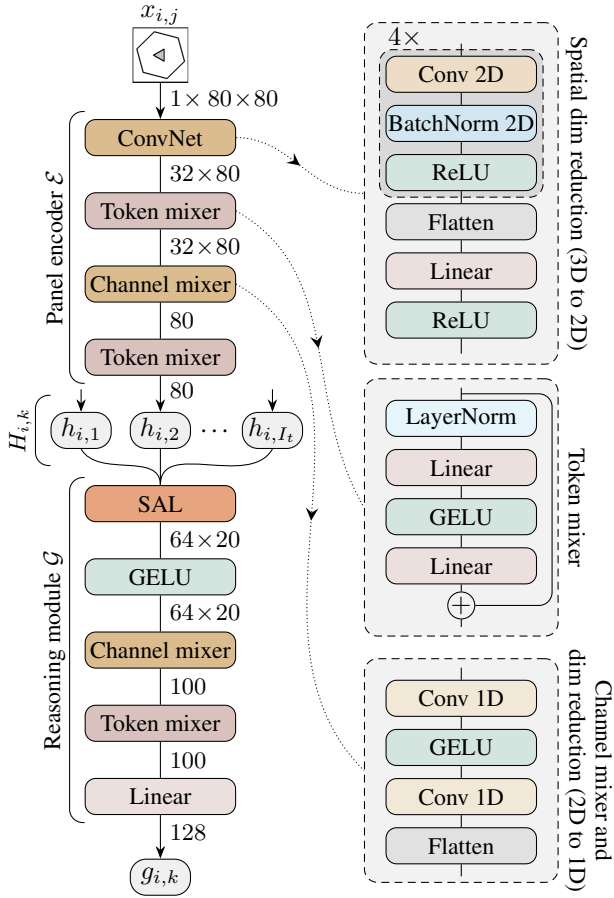
\begin{figure}[b!]
    \centering
    \begin{tikzpicture}[node distance = 2cm, auto]
        \begin{scope}[node distance=1.2*\unit]
            \node (im1) [panel] {\rpm{0}};
            \node [yshift=0.4*\unit] at (im1.north) {$x_{i,j}$};
            \node (conv1) [layer, fill=conv, below=of im1] {};
            \node (feed1) [feedforward, below=of conv1] {};
            \node (conv2) [layer, fill=conv, below=of feed1] {};
            \node (feed2) [feedforward, below=of conv2] {};
            \node (h2) [embedding, below=of feed2] {$h_{i,2}$};
            \node (dfp) [layer, fill=dfp, below=of h2] {};
            \node (gelu) [activation, below=of dfp] {};
            \node (conv3) [layer, fill=conv, below=of gelu] {};
            \node (feed3) [feedforward, below=of conv3] {};
            \node (linear1) [linear, below=of feed3] {};
            \node (g) [embedding, below=of linear1, yshift=-0.25*\unit] {$g_{i,k}$};
        \end{scope}

        \begin{scope}[node distance=0.6*\unit]
            \node (h1) [embedding, left=of h2] {$h_{i,1}$};
            \node (hi) [text centered, inner sep=0, minimum height=1.0*\unit, minimum width=1.0*\unit, right=of h2, xshift=-0.3*\unit] {\ldots};
            \node (hp) [embedding, right=of hi, xshift=-0.3*\unit] {$h_{i,I_t}$};
        \end{scope}

        \node [label] at (conv1.center) {ConvNet};
        \node [label] at (feed1.center) {Token mixer};
        \node [label] at (conv2.center) {Channel mixer};
        \node [label] at (feed2.center) {Token mixer};
        \node [label] at (dfp.center) {SAL};
        \node [label] at (gelu.center) {GELU};
        \node [label] at (conv3.center) {Channel mixer};
        \node [label] at (feed3.center) {Token mixer};
        \node [label] at (linear1.center) {Linear};

        \node [label, align=left, anchor=west] at ($ (im1.south)!0.45!(conv1.north) $) {$1\!\times 80\!\times\!80$};
        \node [label, align=left, anchor=west] at ($ (conv1)!0.5!(feed1) $) {$32\!\times\!80$};
        \node [label, align=left, anchor=west] at ($ (feed1)!0.5!(conv2) $) {$32\!\times\!80$};
        \node [label, align=left, anchor=west] at ($ (conv2)!0.5!(feed2) $) {$80$};
        \node [label, align=left, anchor=west] at ($ (feed2)!0.45!(h2) $) {$80$};
        \node [label, align=left, anchor=west] at ($ (dfp)!0.5!(gelu) $) {$64\!\times\!20$};
        \node [label, align=left, anchor=west] at ($ (gelu)!0.5!(conv3) $) {$64\!\times\!20$};
        \node [label, align=left, anchor=west] at ($ (conv3)!0.5!(feed3) $) {$100$};
        \node [label, align=left, anchor=west] at ($ (feed3)!0.5!(linear1) $) {$100$};
        \node [label, align=left, anchor=west] at ($ (linear1)!0.45!(g) $) {$128$};

        \draw [arrow] (im1.south) -- (conv1.north);
        \draw (conv1.south) -- (feed1.north);
        \draw (feed1.south) -- (conv2.north);
        \draw (conv2.south) -- (feed2.north);

        \draw [arrow] ($ (h1.north) + (0, 0.75*\unit) $) -- (h1.north);
        \draw [arrow] (feed2.south) -- (h2.north);
        \draw [arrow] ($ (hp.north) + (0, 0.75*\unit) $) -- (hp.north);

        \draw (h1.south) to[out=300, in=90] (dfp.north);
        \draw (h2.south) -- (dfp.north);
        \draw (hp.south) to[out=240, in=90] (dfp.north);

        \draw (dfp.south) -- (gelu.north);
        \draw (gelu.south) -- (conv3.north);
        \draw (conv3.south) -- (feed3.north);
        \draw (feed3.south) -- (linear1.north);
        \draw [arrow] (linear1.south) -- (g.north);

        \node (brace_e1) at ($ (conv1.north west) + (-0.5*\unit, 0.25*\unit) $) {};
        \node (brace_e2) at ($ (feed2.south west) + (-0.5*\unit, -0.25*\unit) $) {};
        \draw [rounded corners=5pt] ($ (brace_e1) + (0.5*\unit, 0) $) -- (brace_e1.center) -- (brace_e2.center) -- ($ (brace_e2) + (0.5*\unit, 0) $);
        \node [label, rotate=90, yshift=0.5*\unit] at ($ (brace_e1)!0.5!(brace_e2) $) {Panel encoder $\mathcal{E}$};

        \node (brace_h1) at ($ (h1.north west) + (-0.5*\unit, 0.55*\unit) $) {};
        \node (brace_h2) at ($ (h1.south west) + (-0.5*\unit, -0.55*\unit) $) {};
        \draw [rounded corners=5pt] ($ (brace_h1) + (0.5*\unit, 0) $) -- (brace_h1.center) -- (brace_h2.center) -- ($ (brace_h2) + (0.5*\unit, 0) $);
        \node [label, rotate=90, yshift=0.5*\unit] at ($ (brace_h1)!0.5!(brace_h2) $) {$H_{i,k}$};

        \node (brace_f1) at ($ (dfp.north west) + (-0.5*\unit, 0.25*\unit) $) {};
        \node (brace_f2) at ($ (linear1.south west) + (-0.5*\unit, -0.25*\unit) $) {};
        \draw [rounded corners=5pt] ($ (brace_f1) + (0.5*\unit, 0) $) -- (brace_f1.center) -- (brace_f2.center) -- ($ (brace_f2) + (0.5*\unit, 0) $);
        \node [label, rotate=90, yshift=0.5*\unit] at ($ (brace_f1)!0.5!(brace_f2) $) {Reasoning module $\mathcal{G}$};

        \begin{scope}[node distance=0.4*\unit]
            \node (pe1) [emptylayer] at ($ (im1) + (10.0*\unit, 1.0*\unit) $) {};
            \node (con1) [conv2d, below=of pe1] {};
            \node (bn1) [batchnorm, below=of con1] {};
            \node (relu1) [activation, below=of bn1] {};
            \node (flat1) [layer, below=of relu1] {};
            \node (pelin1) [linear, below=of flat1] {};
            \node (perelu1) [activation, below=of pelin1] {};
            \node (pe2) [emptylayer, below=of perelu1] {};
        \end{scope}
        \node (pe_south) at ($ (perelu1.south) - (0, 0.5*\unit) $) {};
        \node (pe_south_east) at ($ (perelu1.south) - (0, 0.5*\unit) $) {};

        \node [label] at (con1.center) {Conv 2D};
        \node [label] at (bn1.center) {BatchNorm 2D};
        \node [label] at (relu1.center) {ReLU};
        \node [label] at (pelin1.center) {Linear};
        \node [label] at (perelu1.center) {ReLU};
        \node [label] at (flat1.center) {Flatten};

        \draw (pe1.south) -- (con1.north);
        \draw (con1.south) -- (bn1.north);
        \draw (bn1.south) -- (relu1.north);
        \draw (relu1.south) -- (flat1.north);
        \draw (flat1.south) -- (pelin1.north);
        \draw (pelin1.south) -- (perelu1.north);
        \draw (perelu1.south) -- (pe_south.center);

        \begin{scope}[on background layer]
            \node (bg_panel_encoder) [fit={(pe1.south) (con1) (perelu1) ($ (pe2.north) + (0, 0.25*\unit) $)}, background, fill=black!5, inner sep=0.7*\unit] {};
        \end{scope}
        \node [label, rotate=270, yshift=0.6*\unit] at (bg_panel_encoder.east) {Spatial dim reduction (3D to 2D)};

        \begin{scope}[on background layer]
            \node (bg_fourx) [fit={(con1) (relu1)}, background, fill=black!15, inner sep=0.2*\unit] {};
        \end{scope}

        \node (fourx) [align=center, anchor=west, yshift=0.4*\unit] at (bg_fourx.north west) {$4\times$};

        \begin{scope}[node distance=0.4*\unit]
            \node (ln1) [layernorm, below=of pe2] {};
            \node (lin1) [linear, below=of ln1] {};
            \node (gelu1) [activation, below=of lin1] {};
            \node (lin2) [linear, below=of gelu1] {};
            \node (ff2) [emptylayer, below=of lin2] {};
        \end{scope}

        \node (res1) at ($ (pe2)!0.5!(ln1) $) {};
        \node (res2) at ($ (ln1.east |- res1.center) + (0.5*\unit, 0) $) {};
        \node [circle, draw=black, inner sep=0] (res4) at ($ (lin2)!0.5!(ff2) - (0, 0.5*\unit) $) {$+$};
        \node (res3) at ($ (lin2.east |- res4.center) + (0.5*\unit, 0) $) {};

        \node [label] at (ln1.center) {LayerNorm};
        \node [label] at (lin1.center) {Linear};
        \node [label] at (gelu1.center) {GELU};
        \node [label] at (lin2.center) {Linear};

        \draw (pe2.south) -- (ln1.north);
        \draw (ln1.south) -- (lin1);
        \draw (lin1.south) -- (gelu1.north);
        \draw (gelu1.south) -- (lin2.north);
        \draw (lin2.south) -- (res4.north);
        \draw (res4.south) -- ($ (res4.south) - (0, 0.25*\unit) $);

        \draw [rounded corners=5pt] (res1.center) -- (res2.center) -- (res3.center) -- (res4.east);

        \begin{scope}[on background layer]
            \node (bg_feed_forward) [fit={($ (pe2.south) + (0, -0.5*\unit) $) (ln1) (ff2.center)}, background, fill=black!5, inner sep=0.7*\unit] {};
        \end{scope}
        \node [label, rotate=270, yshift=0.6*\unit] at (bg_feed_forward.east) {Token mixer};

        \begin{scope}[node distance=0.4*\unit]
            \node (cm1) [emptylayer, below=of ff2, yshift=0.5*\unit] {};
            \node (c1d1) [conv1d, below=of cm1] {};
            \node (gelu1) [activation, below=of c1d1] {};        
            \node (c1d2) [conv1d, below=of gelu1] {};
            \node (flat2) [layer, below=of c1d2] {};
            \node (cm2) [emptylayer, below=of flat2] {};
        \end{scope}

        \node [label] at (c1d1.center) {Conv 1D};
        \node [label] at (gelu1.center) {GELU};
        \node [label] at (c1d2.center) {Conv 1D};
        \node [label] at (flat2.center) {Flatten};

        \draw (cm1.south) -- (c1d1.north);
        \draw (c1d1.south) -- (gelu1);
        \draw (gelu1.south) -- (c1d2.north);
        \draw (c1d2.south) -- (flat2.north);
        \draw (flat2.south) -- (cm2.north);

        \begin{scope}[on background layer]
            \node (bg_channel_mixer) [fit={($ (cm1.south) + (0, -0.5*\unit) $) (c1d1) ($ (cm2.north) + (0, 0.5*\unit) $)}, background, fill=black!5, inner sep=0.7*\unit] {};
        \end{scope}
        \node [label, rotate=270, yshift=1.0*\unit, align=center] at (bg_channel_mixer.east) {Channel mixer and\\dim reduction (2D to 1D)};

        \begin{scope}[decoration={markings, mark=at position 0.5 with {\arrow[scale=1.5,>=stealth]{>}}}] 
            \draw [postaction={decorate}, densely dotted] (conv1.east) to[out=0, in=180] (bg_panel_encoder.west);
            \draw [postaction={decorate}, densely dotted] (feed1.east) to[out=330, in=120] (bg_feed_forward.west);
            \draw [postaction={decorate}, densely dotted] (conv2.east) to[out=330, in=135] (bg_channel_mixer.west);
        \end{scope}
    \end{tikzpicture}
    \caption{
    \textbf{SCAR architecture.}
    Each input panel $x_{i,j}$ is embedded separately with the panel encoder $\mathcal{E}$ to a latent representation $h_{i,j}$.
    The embeddings are arranged into $A_t$ groups $\{ H_{i,k} \}_{k=1}^{A_t}$.
    The figure depicts processing of a single group $H_{i,k}$ corresponding to answer $k$.
    Vectors belonging to this group are fused and processed with the reasoning module $\mathcal{G}$.
    The resultant representation $g_{i,k}$ is used to predict the index of the correct answer and optionally the encoded rules.
    }
    \label{fig:model}
\end{figure}

\subsection{Model architecture}\label{subsec:model}
We propose the \emph{unified model for solving Single-Choice Abstract visual Reasoning tasks} (SCAR), which is an end-to-end architecture based on SAL, illustrated in Fig.~\ref{fig:model}.
A detailed description of all SCAR hyperparameters is provided in Appendix~\ref{sec:model-details}.

SCAR starts with a panel encoder $\mathcal{E}$ which processes each $x_{i,j}$ with a 4-layer convolutional module to extract low-level image features.
Each layer comprises a 2D $3 \times 3$ convolution, 2D Batch Normalization~\cite{ioffe2015batch}, and the ReLU~\cite{nair2010rectified} activation.
The output of the module is flattened along the spatial dimensions to a 2D matrix and processed with a token mixer module, which fuses feature representations from different spatial regions of the considered panel.
Token mixer is composed of Layer Normalization~\cite{ba2016layer}, two linear layers with the GELU~\cite{hendrycks2016gaussian} nonlinearity in-between, and a residual connection.
Next, a channel mixer module consisting of two 1D $1 \times 1$ convolutions interleaved with GELU is employed to mix features from a given spatial region and flatten the matrix to a 1D vector.
Another token mixer transforms the flattened vector into the panel embedding $h_{i,j}$.

Next, the resultant panel embeddings $\{ h_{i,j} \}_{j=1}^{P_t}$ are arranged into $A_t$ groups $\{ H_{i,k} \}_{k=1}^{A_t}$.
Each group is then processed separately by the reasoning module $\mathcal{G}$, which consists of SAL, GELU, channel mixer, token mixer, and a linear layer that produces $g_{i,k}$.
Lastly, two linear layers are interleaved with GELU to map $g_{i,k}$ onto the predicted score $s_{i,k}$.

\subsection{Unified reasoning}
The proposed model is applied to three challenging AVR problems of different structure --
for RPMs $r = c = 3$,
for VAPs $r = 2$ and $c = 3$,
and for O3 tests $r = 1$ and $c \in \{4, 5, 6\}$.
To enable solving these tree diverse tasks with a common SCAR model, the following design choices have been made.

\paragraph{SCAR adaptation to diverse structures.}
The proposed formulation of dynamic weight adaptation in SAL utilizes an underlying weight matrix $W^*$, for which $R$ and $C$ (its first two dimensions) have to be evenly divisible by $r$ and $c$, see Eq.~\ref{eq:sliding-window}.
To instantiate SCAR, we calculate the least common multiple of all possible values for $r \in \{ 1, 2, 3 \}$ and $c \in \{ 3, 4, 5, 6 \}$, which gives $R = 6$ and $C = 60$.

\paragraph{Solving O3 tests.}
O3 tests have a fundamentally different goal than RPMs and VAPs. Instead of selecting an answer that correctly completes the context grid, in O3 tests it is required to select a panel that differentiates most from the remaining ones.
To handle these opposing settings in one unified model, we cast O3 to the setting where a subset of $P_t - 1$ panels with the most common features has to be chosen.
Putting this into the introduced framework, we arrange the panel embeddings $\{ h_{i,j} \}_{j=1}^{P_t}$ into groups $\{ H_{i,k} \}_{k=1}^{A_t}$ (for O3 tests $P_t = A_t$) such that $H_{i,k} = \{ h_{i,j}\ \vert\ j \neq k \}_{j=1}^{P_t}$.
This allows to interpret $g_{i,k}$ as a latent representation of semantic congruency of the considered subset of panels.

\paragraph{Training objectives.}
To optimize parameters of the model, we compute cross-entropy $\mathcal{L}^{\text{CE}}$ between the predicted distribution $\widehat{p}$ and ground-truth $p$ obtained by one-hot encoding of the target label $k$.
In addition, some considered datasets provide auxiliary metadata referring to the problem instances~\cite{santoro2018measuring}.
In particular, matrices from PGM, I-RAVEN, and VAP provide ground-truth annotations of the underlying matrix rules: $M_i = (X_i, y_i, r_i)$.
To utilize this additional training signal, for each task we employ a separate rule prediction head $\mathcal{R}_t ( \bm{g}_{i,k} ) = \widehat{\bm{r}}_{i,k}$ composed of two linear layers with GELU in-between.
$\mathcal{R}$ predicts the rule representation $r_i$ encoded with sparse encoding~\cite{malkinski2020multi}.
Binary cross-entropy $\mathcal{L}^{\text{AUX}}$ is used to assess the fitness between $\widehat{r}_i$ and $r_i$.
In effect, the model is trained to minimise $\mathcal{L}^{\text{CE}} + \beta\mathcal{L}^{\text{AUX}}$, where $\beta = 10$ is a fixed balancing coefficient.

\paragraph{Mini-batch training.}
In MTL experiments, each training batch is composed of matrices belonging to a single randomly-chosen task.
This ensures that the operation $W^TG_{i,k}$ (Eq.~\ref{eq:dynamic-weight-adaptation-linear}) can be efficiently computed for the whole batch with a single matrix multiplication.

\begin{table}[t]
    \small
    \centering
    \caption{
    \textbf{Single-task learning.}
    Test accuracy of universal (WReN, RNN, SCAR) and task-specific (CoPINet, SRAN, SCL) models in solving a single task.
    Due to architectural constraints, the considered task-specific models created with RPMs in mind can't be directly applied to solve VAPs and O3 tests.
    Best results in each group of models are highlighted in bold and second best are underlined.
    }
    \begin{sc}
        \begin{tabular}{l||ccc|c|c}
            \toprule
            \multirow{2}{*}{Model} & \multicolumn{5}{c}{Test accuracy (\%)} \\
            & \scriptsize G-set & \scriptsize PGM-S & \scriptsize I-RAVEN & \scriptsize VAP-S & \scriptsize O$3$ \\
            \midrule
            \midrule
            WReN & 34.7 & 25.1 & 21.9 & 73.6 & \underline{65.6} \\
            RNN & 49.6 & 29.4 & 38.9 & 88.1 & 51.2 \\
            SCL$+$SAL & \textbf{89.9} & \textbf{51.5} & \textbf{95.5} & \underline{92.6} & 63.1 \\
            SCAR & \underline{82.1} & \underline{46.1} & \underline{94.7} & \textbf{92.9} & \textbf{86.7} \\
            \midrule
            CoPINet & 63.5 & 37.3 & 47.1 & N/A & N/A \\
            SRAN & \underline{82.3} & 39.4 & 66.9 & N/A & N/A \\
            RelBase & 81.3 & \underline{50.3} & \underline{93.7} & N/A & N/A \\
            SCL & \textbf{83.4} & \textbf{51.1} & \textbf{95.5} & N/A & N/A \\
            \bottomrule
        \end{tabular}
    \end{sc}
    \label{tab:stl}
\end{table}

\section{Experiments}\label{sec:experiments}
The experimental evaluation of SCAR focuses on the following three settings:
(1) single-task learning (STL), where the model is trained and evaluated on the same task,
(2) multi-task learning (MTL), where the model is simultaneously trained on several tasks and then fine-tuned, separately for each task, and
(3) transfer learning (TL), where the model is pre-trained on a base set of tasks and fine-tuned on a novel target task.
In each scenario, the performance of SCAR is compared to leading AVR literature models, both task-specific and universal.
The code required to reproduce the experiments is available online\footnote{\url{www.github.com/mikomel/sal}}.
Extended results are provided in Appendix~\ref{sec:extended-results}.

\paragraph{Tasks.}
The experiments are conducted on three challenging AVR problems.
Firstly, we consider RPMs belonging to three datasets:
G-set~\cite{mandziuk2019deepiq} with $1\,100$ matrices;
PGM-S, a subset of the Neutral regime of PGM~\cite{santoro2018measuring} where the training split was obtained by uniformly sampling instances from the parent dataset preserving the original ratio of object--attribute--rule triplets;
and I-RAVEN~\cite{zhang2019raven,hu2021stratified} with $70$K samples.
Secondly, we consider the VAP dataset~\cite{hill2018learning} to construct the VAP-S dataset with training matrices sampled analogously to PGM-S.
Thirdly, the O3 tests~\cite{mandziuk2019deepiq} with $1,000$ instances are utilized.
In G-set, I-RAVEN, and O3 datasets we uniformly allocate $60\%$ / $20\%$ / $20\%$ samples to train / val / test splits, resp.
PGM-S contains $42$K / $20$K / $200$K matrices and VAP-S has $42$K / $10$K / $100$K instances in train / val / test splits, resp.
Altogether, the selection of tasks comprises a diverse set of AVR matrices ranging from visually simple instances from G-set and O3, through more challenging ones from PGM and VAP, up to matrices from I-RAVEN with hierarchical structure.
In addition, the variability of the component dataset sizes allows to evaluate the models' performance in data-constrained scenarios.
At the same time, considering subsets of PGM and VAP allows the construction of a joint dataset for MTL where the size of these initially large datasets doesn't dominate the smaller ones.

\paragraph{Baselines.}
SCAR is compared with WReN~\cite{santoro2018measuring} and RNN~\cite{hill2018learning}, the two alternative universal models, capable of processing AVR tasks with variable structure, which is a prerequisite for their application to multi-task scenarios.
We also consider a variant of SCL~\cite{wu2020scattering} where the first layer in its relationship network $\mathcal{N}^r$ is replaced with SAL (denoted as SCL+SAL).
In addition, we compare SCAR with selected state-of-the-art task-specific models: SRAN~\cite{hu2021stratified}, CoPINet~\cite{zhang2019learning}, RelBase~\cite{spratley2020closer} and SCL.
These additional baselines are included only in the RPM settings, as the architectural constraints prevent their direct application to tasks with different structure.

\paragraph{Experimental setting.}
In the default setting, we use batches of size 32 for G-set and O$3$, and 128 for the remaining datasets.
Early stopping is performed after the model's performance stops improving on a validation set for 17 successive epochs.
We use the Adam~\cite{kingma2014adam} optimizer with $\beta_1=0.9$, $\beta_2=0.999$ and $\epsilon=10^{-8}$, the learning rate is initialized to $0.001$ and reduced $10$-fold if the validation loss doesn't improve for 5 subsequent epochs.
In each experiment, data augmentation is employed with $50\%$ probability of being applied to a particular sample.
When applied, a pipeline of transformations (vertical flip, horizontal flip, rotation by 90 degrees, rotation, transposition) is constructed, each with $25\%$ probability of being adopted.
The resultant pipeline is applied to each image in the matrix in the same way.
Each training job was run on a node with a single NVIDIA DGX A100 GPU.

\subsection{Single-task learning}
Table~\ref{tab:stl} presents STL results, where each model is trained and evaluated on the same dataset.
Among universal methods, either SCL+SAL or SCAR achieved the best results on each considered dataset, demonstrating SAL's applicability to solving AVR tasks with diverse structures, and high efficacy even in tasks with limited number of training samples (G-set and O$3$).
When compared to task-specific models, SCAR performed better than two selected baselines (CoPINet and SRAN) and was slightly inferior to SCL.
In contrast to the universal methods, the dedicated RPM models cannot be evaluated on VAPs and O3 tests due to their architectural design.

\addtolength{\tabcolsep}{-2pt}
\begin{table*}[t]
    \small
    \centering
    \caption{
    \textbf{Multi-task learning.}
    Test accuracy of universal and task-specific models pre-trained simultaneously on multiple tasks and then separately fine-tuned on each task.
    The results shown in the left part of the table were achieved by pre-training only on RPMs (G-set, PGM-S, and I-RAVEN), while those in the right part additionally included VAPs in the pre-training set.
    Values in parentheses show the difference w.r.t. the STL setup (cf. Table~\ref{tab:stl}).
    }
    \begin{sc}
        \resizebox{\textwidth}{!}{\begin{tabular}{l||ccc||ccc|c}
            \toprule
            \multirow{3}{*}{Model} & \multicolumn{7}{c}{Test accuracy (\%)} \\
            & \multicolumn{3}{c||}{Pre-trained on RPMs} & \multicolumn{4}{c}{Pre-trained on RPMs and VAPs} \\
            & \scriptsize G-set & \scriptsize PGM-S & \scriptsize I-RAVEN & \scriptsize G-set & \scriptsize PGM-S & \scriptsize I-RAVEN & \scriptsize VAP-S \\
            \midrule
            \midrule
            WReN~\cite{santoro2018measuring} & 41.6 \scriptsize(+\ 6.9) & 24.5 \scriptsize(-\ 0.6) & 19.4 \scriptsize(-\ 2.5) & 48.8 \scriptsize(+14.1) & 26.1 \scriptsize(+\ 1.0) & 20.2 \scriptsize(-\ 1.7) & 80.6 \scriptsize(+\ 7.0) \\
            RNN~\cite{hill2018learning} & 66.4 \scriptsize(+16.8) & 27.8 \scriptsize(-\ 1.6) & 34.0 \scriptsize(-\ 4.9) & 61.9 \scriptsize(+12.3) & 26.5 \scriptsize(-\ 2.9) & 27.8 \scriptsize(-11.1) & 88.6 \scriptsize(+\ 0.5) \\
            SCL+SAL & \textbf{94.9} \scriptsize(+\ 5.0) & \textbf{58.1} \scriptsize(+\ 6.6) & \textbf{95.4} \scriptsize(-\ 0.1) & \textbf{94.8} \scriptsize(+\ 4.9) & \textbf{68.6} \scriptsize(+17.1) & \textbf{95.3} \scriptsize(-\ 0.2) & \underline{92.8} \scriptsize(+\ 0.2) \\
            SCAR & \underline{91.0} \scriptsize(+\ 8.9) & \underline{53.7} \scriptsize(+\ 7.6) & \underline{94.3} \scriptsize(-\ 0.4) & \underline{93.2} \scriptsize(+11.1) & \underline{54.8} \scriptsize(+\ 8.7) & \underline{94.7} \scriptsize(+\ 0.0) & \textbf{93.0} \scriptsize(+\ 0.1) \\
            \midrule
            CoPINet~\cite{zhang2019learning} & 74.9 \scriptsize(+11.4) & 35.2 \scriptsize(-\ 2.1) & 50.3 \scriptsize(+\ 3.2) & N/A & N/A & N/A & N/A \\
            SRAN~\cite{hu2021stratified} & 87.2 \scriptsize(+\ 4.9) & 40.7 \scriptsize(+\ 1.3) & 68.3 \scriptsize(+\ 1.4) & N/A & N/A & N/A & N/A \\
            SCL~\cite{wu2020scattering} & \textbf{96.2} \scriptsize(+12.8) & \textbf{69.6} \scriptsize(+18.5) & \textbf{95.3} \scriptsize(-\ 0.2) & N/A & N/A & N/A & N/A \\
            RelBase~\cite{spratley2020closer} & \underline{95.7} \scriptsize(+14.4) & \underline{64.0} \scriptsize(+13.7) & \underline{89.3} \scriptsize(-\ 4.4) & N/A & N/A & N/A & N/A \\
            \bottomrule
        \end{tabular}}
    \end{sc}
    \label{tab:mtl}
\end{table*}
\addtolength{\tabcolsep}{2pt}

\begin{table}[t]
    \small
    \centering
    \caption{
    \textbf{Transfer learning.}
    Test accuracy of universal models after transferring knowledge from a set of pre-training tasks to a novel target task.
    R and V denote RPMs and VAPs, resp.
    Values in parentheses denote p.p. change w.r.t. STL (cf. Table~\ref{tab:stl}).
    }
    \begin{sc}
        \begin{tabular}{l||c|cc}
            \toprule
            \multirow{2}{*}{Model} & \multicolumn{3}{c}{Test accuracy (\%)} \\
            & \scriptsize R $\to$ V & \scriptsize R $\to$ O$3$ & \scriptsize R+V $\to$ O$3$ \\
            \midrule
            \midrule
            WReN & 83.3 \scriptsize (+9.7) & \underline{65.4} \scriptsize (-0.2) & 65.5 \scriptsize (-0.1) \\
            RNN & 88.1 \scriptsize (+0.0) & 55.9 \scriptsize (+4.7) & 55.3 \scriptsize (+4.1) \\
            SCL+SAL & \textbf{93.6} \scriptsize (+1.0) & \underline{65.4} \scriptsize (+2.3) & \underline{75.5} \scriptsize (+12.4) \\
            SCAR & \underline{93.0} \scriptsize (+0.1) & \textbf{88.6} \scriptsize (+1.9) & \textbf{89.0} \scriptsize (+2.3) \\
            \bottomrule
        \end{tabular}
    \end{sc}
    \label{tab:tl}
\end{table}

\addtolength{\tabcolsep}{-3pt}
\begin{table}[t]
    \small
    \centering
    \caption{
    \textbf{Ablation study.}
    Test accuracy of SCAR variants where SAL is replaced with RN and LSTM, respectively.
    Both models were trained in the STL setup.
    Values in parentheses denote p.p. change w.r.t. the STL training of the default variant of SCAR with SAL -- see Table~\ref{tab:stl}.
    }
    \begin{sc}
        \resizebox{0.47\textwidth}{!}{\begin{tabular}{l||ccc|c|c}
            \toprule
            & \multicolumn{5}{c}{Test accuracy (\%)} \\
            & \scriptsize G-set & \scriptsize PGM-S & \scriptsize I-RAVEN & \scriptsize VAP-S & \scriptsize O$3$ \\
            \midrule
            \midrule
            RN & 36.5 \scriptsize (-45.6) & 33.2 \scriptsize (-12.9) & 56.1 \scriptsize (-38.6) & 90.5 \scriptsize (-2.4) & 69.5 \scriptsize (-17.2) \\
            LSTM & 43.5 \scriptsize (-38.6) & 26.0 \scriptsize (-20.1) &  41.7 \scriptsize (-53.0) & 90.7 \scriptsize (-2.2) & 52.4 \scriptsize (-34.3) \\
            \bottomrule
        \end{tabular}}
    \end{sc}
    \label{tab:ablation-study}
\end{table}
\addtolength{\tabcolsep}{3pt}

\subsection{Multi-task learning}
Next, we evaluate knowledge reuse capabilities of the adopted methods.
Toward this end, in the MTL setting, the models are first pre-trained on a base set of tasks and then separately fine-tuned on each task.
Two settings for the selection of tasks are considered:
(1) RPMs only, which allows measuring knowledge reuse within a single problem;
and (2) RPMs combined with VAPs, to evaluate knowledge reuse between problems.
In the pre-training phase batches of size 128 are used.

The results are presented in Table~\ref{tab:mtl}.
Among universal models, either SCAR or SCL+SAL achieved the highest scores.
In both settings MTL helped to increase their performance on the most challenging tasks, i.e. G-set and PGM-S, showing SAL's ability to effectively reuse the gained knowledge.
On I-RAVEN and VAP-S their results are on-par with these of STL, however, the models achieved high accuracy already in the STL setup, leaving small room for potential improvement.
For the two other universal methods, MTL proved beneficial in some settings (G-set and VAP-S), while it hindered the performance in other cases (I-RAVEN).
This, contrary to SCAR and SCL+SAL, suggests insufficient capacity of WReN and RNN to leverage previously acquired knowledge.
Regarding the task-specific methods, MTL improved their performance in most experiments, especially on the data-constrained G-set.
Due to architectural constraints, the VAP dataset could not be included in the experiments concerning the RPM-specialised methods.
Generally, adopting MTL approach to solving AVR tasks, proposed in this paper, has proven effective, especially for the data-limited tasks.
At the same time, the results show that MTL need to be designed with care, as in certain settings its application may deteriorate the results, a problem known in the literature as the \emph{negative transfer} effect~\cite{pan2010survey,cao2018partial}.

\subsection{Transfer learning}
To further study knowledge reuse capabilities across different AVR problems, we conduct TL experiments where the models are pre-trained on a set of tasks, and then fine-tuned on a novel target task.
Three such settings are considered: taking all three RPM datasets as pre-training tasks and fine-tuning on VAP-S and O$3$, respectively, and pre-training on RPMs and VAP-S and fine-tuning on O$3$.
Table~\ref{tab:tl} showcases the results.
TL allowed to significantly increase the performance of WReN on VAP-S and improved the results of RNN, SCL+SAL and SCAR on O$3$.
The results amplify the importance of developing universal methods in the AVR domain capable of processing tasks with diverse structures.

\subsection{Ablation study}
Lastly, we perform an ablation study to validate the usefulness of the proposed SAL in the SCAR architecture.
To this end, we develop two variants of SCAR in which SAL is replaced with RN~\cite{santoro2017simple} and LSTM~\cite{hochreiter1997long}, resp., and evaluate them in the STL setting.
The results are presented in Table~\ref{tab:ablation-study}.
First, we observe that both alternatives are competitive to other methods trained with STL (see Table~\ref{tab:stl}).
For instance, the RN variant achieved better results than universal WReN and RNN on G-set, PGM-S, VAP-S, and O$3$, and outperformed task-specific CoPINet on I-RAVEN.
Still, both variants perform significantly worse than the default version of SCAR, which confirms the critical importance of SAL in the proposed model.

\section{Conclusion}\label{sec:conclusion}
In this work we consider the problem of solving diverse AVR tasks with a unified model architecture.
To this end we propose SCAR, a DL model capable of solving a variety of single-choice AVR problems.
SCAR design principles differ from the vast majority of the contemporary AVR models, which are developed with a particular task in mind.
The core idea of the proposed model is SAL, a novel structure-aware layer which adapts its weights to the structure of the considered AVR instance.
In the experiments conducted on RPMs, VAPs, and O$3$ tests, SAL-based models significantly surpassed the performance of other universal methods and even of some task-specific models.
At the same time, the models demonstrated effective knowledge reuse capabilities in MTL and TL settings.
On a more general note, with this work we aim to encourage AVR researchers to shift the research focus from task-specific methods towards universal AVR solvers applicable to a wide selection of AVR tasks.


\section*{Acknowledgements}
This research was carried out with the support of the Laboratory of Bioinformatics and Computational Genomics and the High Performance Computing Center of the Faculty of Mathematics and Information Science Warsaw University of Technology.

{
\small
\bibliography{main}
}

\newpage
\appendix
\onecolumn

\section{Extended results}
\label{sec:extended-results}

Below we describe supplementary results not included in the main paper.
All experiments were run using the same experimental setting as described in the main paper.

\paragraph{RelBase.}
We evaluated RelBase~\cite{spratley2020closer} on VAP-S by adjusting the number of input channels in the convolution layer of the sequence encoder module from 9 to 6, so that it corresponds to the number of context panels in VAP matrices.
In STL, the model achieved $91.4\%$ test accuracy, which is worse that SCL+SAL and SCAR.

\paragraph{STSN.}
We also evaluated Slot Transformer Scoring Network (STSN)~\cite{mondal2023learning}.
The model achieved $24.5\%$, $22.9\%$, and $52.6\%$ test accuracy in STL, and $86.7\%$ ($+62.2$ p.p.), $26.3\%$ ($+3.4$ p.p.), and $53.9\%$ ($+1.3$ p.p.) in MTL, on G-set, PGM-S, and I-RAVEN, resp.
The results are worse than reported by~\citet{mondal2023learning} (e.g. on I-RAVEN), which could be attributed to the difference in training hyperparameters (e.g., the learning rate scheduler or data augmentation methods).

\paragraph{SCL with weight cropping.}
We implemented a variant of SCL that adapts its computation to the considered matrix by selecting a submatrix of its weights based on the number of panels in the matrix.
Namely, the \textit{relationship network} 
$\mathcal{N}^r$ in SCL starts with a linear layer with the weight matrix $W \in \mathbb{R} ^ {9 \times 64}$, where $9$ corresponds to the number of context panels in RPMs (defined in the paper as $I_t$), and $64$ is a manually selected hyperparameter defining the number of output neurons of the layer.
For tasks $t'$ with $I_{t'} < 9$, we use a submatrix of $W$ for computation, i.e., $W := W[\ :I_{t'},\ :\ ]$.
In STL, the model achieved $93.4\%$ on VAP-S ($+0.6$ p.p. compared to the best model) and $66.2\%$ on O3 ($-20.5$ p.p.).
In MTL when pre-trained on RPMs and VAPs, the model achieved $92.8\%$ ($+9.4$ p.p. compared to SCL trained with STL), $68.1\%$ ($+17.0$ p.p.), $95.4\%$ ($-0.1$ p.p.), and $92.8\%$ ($-0.6$ p.p.) on G-set, PGM-S, I-RAVEN, and VAP-S, resp.
In TL, the model achieved $93.4\%$ ($+0.0$ p.p.), $79.6\%$ ($+13.4$ p.p.), and $81.3\%$ ($+15.1$ p.p.) in R $\to$ V, R $\to$ O3, and R+V $\to$ O3, resp.
In summary, while the model is competitive with SCL and SCL+SAL in some setups, it's inferior to SCAR in O3 where knowledge reuse is critical due to the low sample size of the dataset.

\section{Example matrices}\label{sec:examples}
\begin{figure*}[h]
    \centering
    \subfloat[\texttt{Up-Down}]{\includegraphics[width=.3\textwidth]{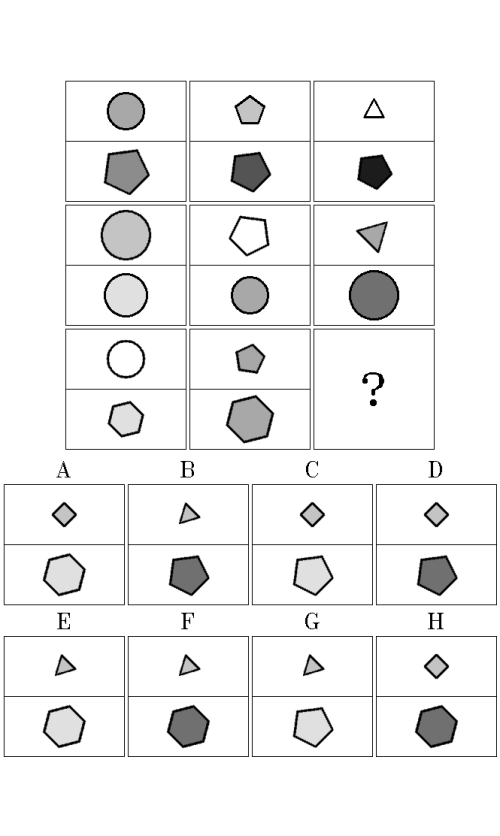}\label{fig:rpm-iraven-1}}
    \hfill
    \subfloat[\texttt{Out-InCenter}]{\includegraphics[width=.3\textwidth]{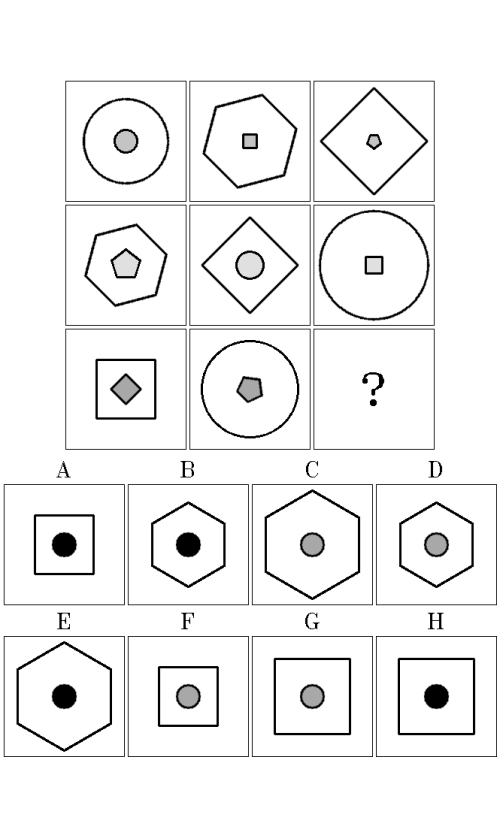}\label{fig:rpm-iraven-2}}
    \hfill
    \subfloat[\texttt{Out-InGrid}]{\includegraphics[width=.3\textwidth]{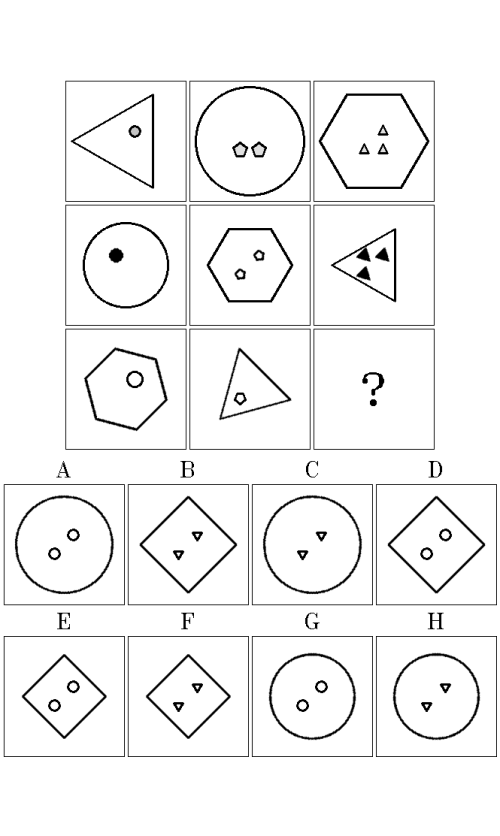}\label{fig:rpm-iraven-3}}
    \caption{\textbf{RPMs from I-RAVEN~\cite{zhang2019raven,hu2021stratified}.}
    The $3 \times 3$ grid of context images has to be completed with the appropriate answer panel (A -- H).
    In I-RAVEN each matrix belongs to one out of seven possible visual configurations.
    Examples of the three selected configurations (\texttt{Up-Down}, \texttt{Out-InCenter}, \texttt{Out-InGrid}) are presented in the figure.
    Each problem instance has up to 8 rules, which can be applied separately to each matrix hierarchy.
    The matrices are governed by the following rules:
    (a) in each row there is a circle, a pentagon, and a triangle in the upper part of exactly one image, while the lower parts of the images in each row contain the same shape with a gradually darker colour from left to right;
    (b) in each row there is a circle, a square, and a hexagon as the outer shape with a gradually increasing size from left to right, and a circle, a square, and a pentagon in the same colour as the inner shape, resp.;
    (c) in each row there is a circle, a triangle, and a hexagon with the same size as the outer shape, and circles, pentagons, and triangles as the inner shapes, resp., whose count in the third column is the sum of the shape counts in the preceding columns.
    The correct answers are F, C, and H, respectively.}
    \label{fig:rpm-iraven}
\end{figure*}

\begin{figure*}[t]
    \centering
    \subfloat[]{\includegraphics[width=.3\textwidth]{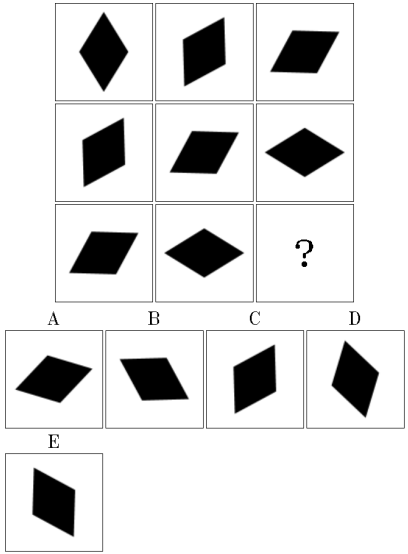}\label{fig:rpm-deepiq-1}}
    \hfill
    \subfloat[]{\includegraphics[width=.3\textwidth]{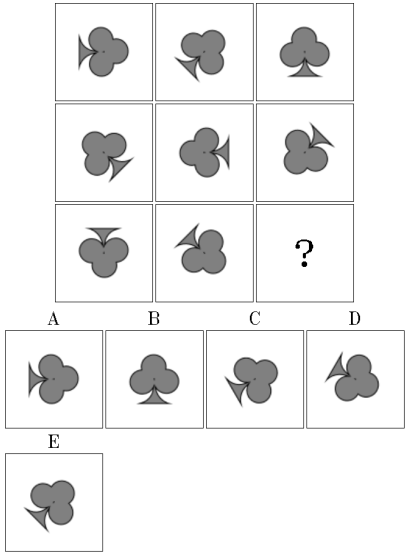}\label{fig:rpm-deepiq-2}}
    \hfill
    \subfloat[]{\includegraphics[width=.3\textwidth]{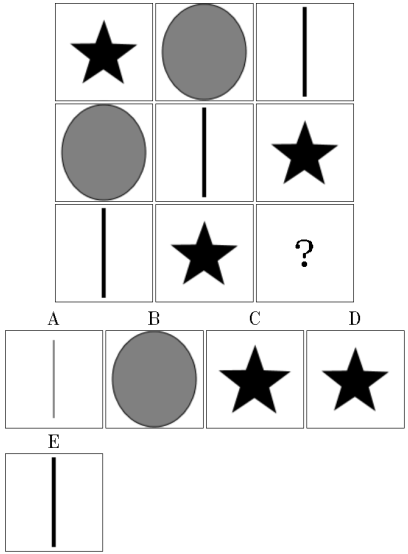}\label{fig:rpm-deepiq-3}}
    \caption{\textbf{RPMs from G-set~\cite{mandziuk2019deepiq}.}
    The $3 \times 3$ grid of context images has to be completed with the appropriate answer panel (A -- E).
    The matrices are governed by the following rules:
    (a) and (b) progression applied to object rotation;
    (c) in each row there are the same three object shapes.
    The correct answers are B, A, and B, respectively.}
    \label{fig:rpm-gset}
\end{figure*}

\begin{figure*}[t]
    \centering
    \subfloat[]{\includegraphics[width=.3\textwidth]{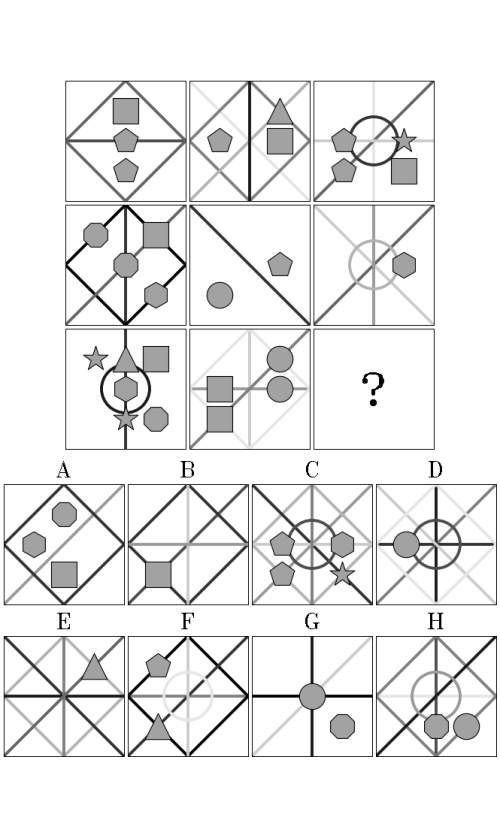}\label{fig:rpm-pgm-1}}
    \hfill
    \subfloat[]{\includegraphics[width=.3\textwidth]{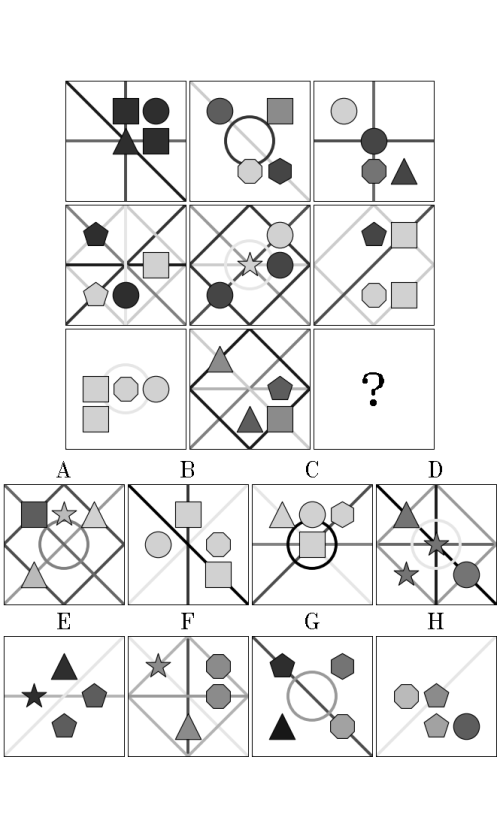}\label{fig:rpm-pgm-2}}
    \hfill
    \subfloat[]{\includegraphics[width=.3\textwidth]{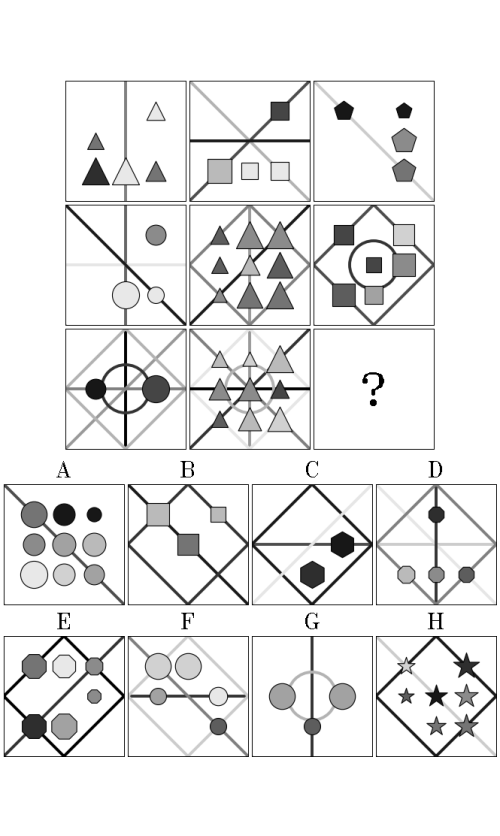}\label{fig:rpm-pgm-3}}
    \caption{\textbf{RPMs from PGM~\cite{santoro2018measuring}.}
    The $3 \times 3$ grid of context images has to be completed with the appropriate answer panel (A -- H).
    The problem instances were sampled from the Neutral PGM regime.
    The matrices are governed by the following rules:
    (a) \texttt{OR} applied column-wise to shape position;
    (b) \texttt{XOR} applied column-wise to shape color;
    (c) progression applied row-wise to shape type.
    The correct answers are C, D, and B, respectively.
    What is more, the examples present the importance of distracting features in PGM matrices (e.g. lines in the example (a)).
    These features are not governed by any rule, and their sole role is to increase the difficulty of the task by distracting the test solver.
    }
    \label{fig:rpm-pgm}
\end{figure*}

\begin{figure*}[t]
    \centering
    \subfloat[]{\includegraphics[width=.3\textwidth]{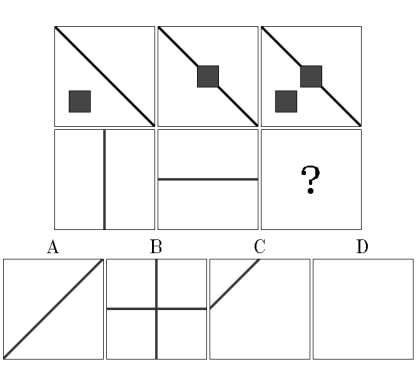}\label{fig:vap-1}}
    \hfill
    \subfloat[]{\includegraphics[width=.3\textwidth]{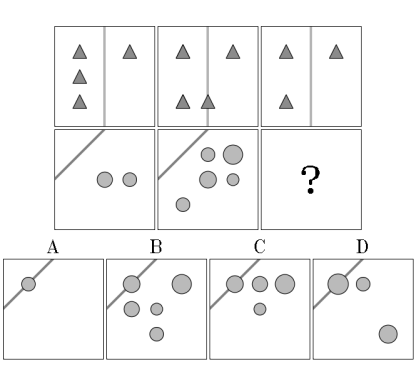}\label{fig:vap-2}}
    \hfill
    \subfloat[]{\includegraphics[width=.3\textwidth]{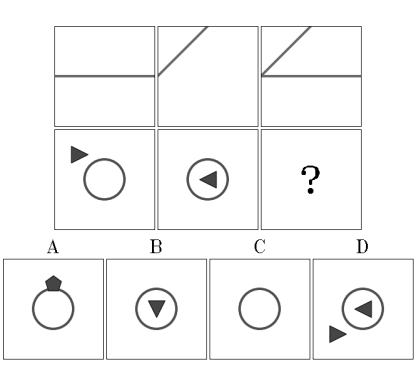}\label{fig:vap-3}}
    \caption{\textbf{VAPs~\cite{hill2018learning}.}
    The $2 \times 3$ grid of context images has to be completed with the appropriate answer panel (A -- D).
    The problem instances were sampled from the Novel Domain Transfer regime, in which a relation presented in a source domain (first row) has to be abstracted and applied to a novel target domain (second row).
    The matrices are governed by the following rules:
    (a) \texttt{XOR} abstracted from shape position to line type;
    (b) \texttt{AND} abstracted from shape position to shape size;
    (c) \texttt{XOR} abstracted from line type to shape type.
    The correct answers are B, A, and D, respectively.
    Similarly to the matrices from PGM, the sole role of certain elements (e.g. the background lines and the circles, respectively in (b) and (c)) is to serve as distractors, to increase the difficulty of the task.}
    \label{fig:vap}
\end{figure*}

\begin{figure*}[t]
    \centering
    \subfloat[]{\includegraphics[width=.4\textwidth]{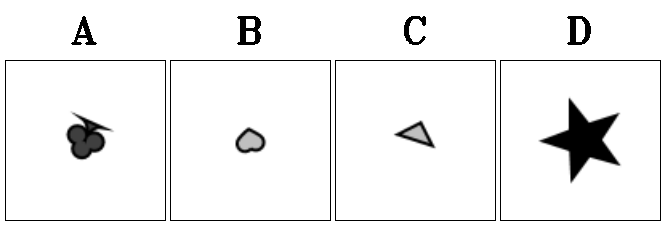}\label{fig:o3-1}}
    \hfill
    \subfloat[]{\includegraphics[width=.5\textwidth]{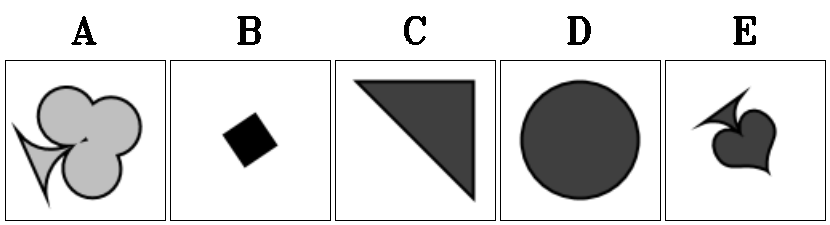}\label{fig:o3-2}}
    \caption{\textbf{O3 tests~\cite{mandziuk2019deepiq}.}
    An odd panel has to be identified among the provided images (A -- D / E).
    The matrices are governed by the following rules:
    (a) all but one shapes have small size;
    (b) all but one shapes have dark grey colour.
    The correct answers are D, and A, respectively.}
    \label{fig:o3}
\end{figure*}

Examples of RPM, VAP, and O3 matrices from the datasets considered in the paper are presented in Figs.~\ref{fig:rpm-iraven}--~\ref{fig:o3}.

\section{Model architecture details}\label{sec:model-details}
Hyperparameters of the considered models are listed in Table~\ref{tab:hyperparameters}.

\vfill

\vfill

\begin{table}[t]
    \small
    \centering
    \caption{
    \textbf{Hyperparameters.}
    The hyperparameters of $\mathcal{E}$ and $\mathcal{G}$ are specific to SCAR, while those of $\mathcal{D}$ and $\mathcal{R}$ are shared by all considered models.
    The parameters
    of convolution layers are denoted as $[$\#~input channels $\to$ \#~output channels, kernel size, stride, padding$]$;
    of linear layers as $[$\#~input neurons $\to$ \#~output neurons$]$;
    of token mixers as $[$\#~input neurons, \#~hidden neurons, \#~output neurons$]$;
    of channel mixers as $[$1st conv layer$]$ -- $[$2nd conv layer$]$ using the same notation as for convolution layers;
    of SAL as $[R, C, d_h, L, d_v]$.
    The number of neurons in BatchNormalization (BN) layers is determined by the number of output channels in the preceding layer.
    The number of output neurons $\vert r \vert$ in the last layer of $\mathcal{R}$ is determined by the number of unique rules for a given dataset (50 for PGM, 40 for I-RAVEN, and 28 for VAP).
    }
    \begin{sc}
        \begin{tabular}{l|c}
            \toprule
            Layer & Hyperparameters \\
            \midrule

            \midrule
            \multicolumn{2}{l}{Panel encoder $\mathcal{E}$} \\
            \midrule
            Conv2d-BN-ReLU & $[1 \to 16, 3 \times 3, 2 \times 2, 1]$ \\
            Conv2d-BN-ReLU & $[16 \to 16, 3 \times 3, 1 \times 1, 1]$ \\
            Conv2d-BN-ReLU & $[16 \to 32, 3 \times 3, 1 \times 1, 1]$ \\
            Conv2d-BN-ReLU & $[32 \to 32, 3 \times 3, 1 \times 1, 1]$ \\
            Flatten & \\
            Linear-ReLU & $[1600 \to 80]$ \\
            Token mixer & $[80 \to 320 \to 80]$ \\
            Channel mixer & $[32 \to 128, 8, 8, 0]$ -- $[128 \to 8, 1, 1, 0]$ \\
            Token mixer & $[80 \to 320 \to 80]$ \\

            \midrule
            \multicolumn{2}{l}{Reasoning module $\mathcal{G}$} \\
            \midrule
            SAL-ReLU & $[6, 60, 80, 20, 64]$ \\
            Channel mixer & $[64 \to 32, 1, 1, 0]$ -- $[32 \to 5, 1, 1, 0]$ \\
            Flatten & \\
            Token mixer & $[100 \to 400 \to 100]$ \\
            Linear & $[100 \to 128]$ \\

            \midrule
            \multicolumn{2}{l}{Alignment score decoder $\mathcal{D}$} \\
            \midrule
            Linear-GELU & $[128 \to 128]$ \\
            Linear & $[128 \to 1]$ \\

            \midrule
            \multicolumn{2}{l}{Rule prediction head $\mathcal{R}$} \\
            \midrule
            Linear-GELU & $[128 \to 128]$ \\
            Linear & $[128 \to \vert r \vert]$ \\

            \bottomrule
        \end{tabular}
    \end{sc}
    \label{tab:hyperparameters}
    \vspace{9.8cm}
\end{table}

\end{document}